\documentclass[10pt,twocolumn,letterpaper]{article}
\usepackage{cvpr}              
\usepackage{multirow}
\usepackage{float}
\usepackage{amsmath}
\usepackage{amsfonts}
\usepackage{multirow}
\usepackage{xcolor}
\usepackage{graphicx}
\usepackage{amsmath}
\usepackage{amssymb}
\usepackage{booktabs}
\usepackage{makecell}
\usepackage{tabularx}
\usepackage{rotating}
\usepackage[normalem]{ulem}
\usepackage[title]{appendix}
\definecolor{cvprblue}{rgb}{0.21,0.49,0.74}
\usepackage[pagebackref,breaklinks,colorlinks,allcolors=cvprblue]{hyperref}

\usepackage[capitalize]{cleveref}
\crefname{section}{Sec.}{Secs.}
\Crefname{section}{Section}{Sections}
\Crefname{table}{Table}{Tables}
\crefname{table}{Tab.}{Tabs.}
\usepackage[accsupp]{axessibility}

\begin{document}

\title{Flow-NeRF: Joint Learning of Geometry, Poses, and Dense Flow within Unified Neural Representations}

\author{
\begin{tabular}[t]{@{}c@{}}
Xunzhi Zheng \quad Dan Xu 
\end{tabular}\\[1ex]
\begin{tabular}[t]{@{}c@{}}
Department of Computer Science and Engineering, HKUST
\end{tabular}\\[0.5ex]
{\tt\small \{xzhengba, danxu\}@cse.ust.hk}
}

\maketitle

\begin{abstract}
Learning accurate scene reconstruction without pose priors in neural radiance fields is challenging due to inherent geometric ambiguity. Recent development either relies on correspondence priors for regularization or uses off-the-shelf flow estimators to derive analytical poses. However, the potential for jointly learning scene geometry, camera poses, and dense flow within a unified neural representation remains largely unexplored. In this paper, we present \textbf{Flow-NeRF}, a unified framework that simultaneously optimizes scene geometry, camera poses, and dense optical flow all on-the-fly. To enable the learning of dense flow within the neural radiance field, we design and build a bijective mapping for flow estimation, conditioned on pose. To make the scene reconstruction benefit from the flow estimation, we develop an effective feature enhancement mechanism to pass canonical space features to world space representations, significantly enhancing scene geometry. We validate our model across four important tasks, i.e., novel view synthesis, depth estimation, camera pose prediction, and dense optical flow estimation, using several datasets. Our approach surpasses previous methods in almost all metrics for novel-view view synthesis and depth estimation and yields both qualitatively sound and quantitatively accurate novel-view flow. Our project page is https://zhengxunzhi.github.io/flownerf/.
\end{abstract}    
\vspace{-10pt}
\section{Introduction}
\label{sec:intro}
Neural Radiance Fields (NeRF)~\cite{mildenhall2021nerf, 2021Mip, barron2023zipnerf, mueller2022instant, Yu2022MonoSDF} enable photo-realistic novel view synthesis and scene reconstruction; however, most methods require camera poses obtained from traditional Structure-from-Motion (SfM) pipelines, such as COLMAP~\cite{schonberger2016structure}, as input for training. Nevertheless, the pose initialization can fail in various cases, e.g., in textureless scenes. Recent studies on pose-free NeRF~\cite{bian2023nope,lin2021barf} and NeRF-based SLAM~\cite{zhu2022nice, zhu2023nicer, li2023dense} have shown that poses can be optimized in conjunction with scene geometry. 
However, optimizing all these objectives can be challenging due to the lack of long-range cross-frame consistency constraints, thus leading to poor reconstruction quality.

\begin{figure}[!t]
\centering \includegraphics[width=0.476\textwidth]{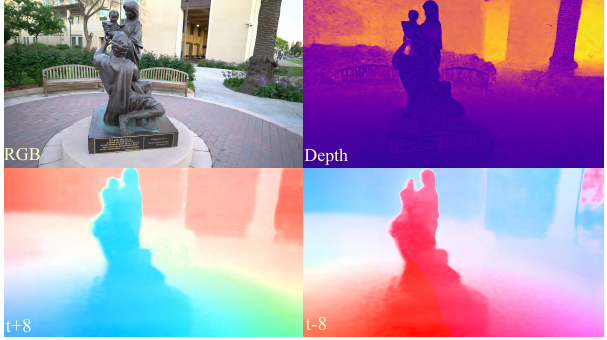}
\vspace{-15pt}
\caption{{Our Flow-NeRF model can simultaneously infer novel-view image, novel-view depth, and long-range novel-view flow \emph{without requiring pose prior}}. While we train the model solely on consecutive forward pseudo flow, it is capable of inferring both forward and backward long-range novel-view flows that are plausible (as illustrated in the two bottom images). In this figure, t+8 and t-8 denote novel-view forward and backward flow, respectively, with a frame interval of 8.}
\vspace{-18pt}
\label{fig:teaser}
\end{figure}

Recently, several NeRF methods~\cite{zhu2023nicer, meuleman2023progressively, smith2023flowcam, smith2024flowmap} have utilized flow supervision to constrain the learning of geometry and cross-view consistency, demonstrating the clear benefits that optical flow can provide for NeRF optimization.
However, these methods treat flow solely as a regularizer for camera pose optimization, leaving its potential to enhance novel-view synthesis and scene geometry largely unexplored.
We demonstrate for the first time that dense flow can serve as a simultaneous optimization goal within an unposed NeRF framework.
This joint optimization can enhance both novel-view synthesis and scene geometry by leveraging the inherent dense cross-view correspondences constrained by flow during optimization.
This is accomplished through the joint optimization of appearance, geometry, camera poses, and dense flow by enforcing the different objectives to learn unified underlying implicit scene representations that couple all the optimization objectives.

On the other hand, many downstream applications (e.g., localization) require accurate long-range correspondences as input. However, these correspondences cannot be inferred from the vanilla NeRF. We argue that for more comprehensive scene modeling, the model should be capable of inferring correspondences between different frames. We therefore target a new objective, i.e., \emph{novel-view flow}, to model the dense correspondences between novel views within the NeRF framework. We enable our model to infer accurate long-range flow (see Fig.~\ref{fig:teaser}) between arbitrary novel views through a novel pose-conditioned input design.

\par To address the previously discussed problems, we propose \textbf{Flow-NeRF}, a unified framework that jointly learns the flow and the geometry fields from two branches. It leverages one key observation: both fields should share underlying scene representations as they inherently model the same physical 3D scene. \emph{Firstly}, we propose a shared point sampling mechanism to forward the same points with alternative representations to these two branches. 
\emph{Secondly}, we propose to use the learned camera pose as a frame identifier to condition the bijective mapping network of the flow branch. This interesting design enables pose-conditioned novel-view flow estimation (see Fig.~\ref{fig:teaser}). 
\emph{Thirdly}, we propose an effective feature message-passing strategy to enhance the feature representation of the geometry branch by incorporating feature messages from the flow branch. This unique design largely improves the results of novel view synthesis and depth prediction. We verify our model on 3 datasets: {Tanks and Temple}~\cite{knapitsch2017tanks}, 
{ScanNet}~\cite{dai2017scannet}, 
and {Sintel}~\cite{Butler:ECCV:2012}. 
We demonstrate an average PSNR improvement of over 2 points for the Tanks and Temples dataset and 0.8 points for the ScanNet dataset.
Additionally, we achieve significant performance gains across all depth metrics for the ScanNet dataset, indicating that our model effectively enhances geometry. We also conduct a quantitative evaluation of \emph{novel-view flow} prediction on the Sintel dataset. Our method largely outperforms one of the state-of-the-art flow predictors, i.e., RAFT~\cite{ranftl2021vision}, on long-range novel-view flow prediction. We obtain an average endpoint error (EPE) of 1.683 compared to RAFT's 2.089 when the frame interval is increased to 16, despite RAFT being pre-trained on a large dataset and fine-tuned for the target dataset.
In summary, our contributions are as follows: 
\begin{itemize}
\item We propose a novel neural scene representation framework to simultaneously learn novel-view synthesis, camera poses, scene geometry, and dense optical flow. 
\item As far as we know, we are the first work that defines and infers novel-view flows. 
The inference of novel-view flow can provide effective cross-view correspondences from unseen views, which helps achieve holistic scene modeling under a unified NeRF framework. 
\item We improve the modeling of scene appearance and geometry by a large margin through the designed implicit distillation from the flow estimation branch. 
\end{itemize}

\section{Related works}
\label{sec:related}

\par\noindent\textbf{NeRF with Unknown Poses.} Recent methods on pose-unknown NeRF~\cite{bian2023nope, zhu2023nicer, lin2021barf, gao2023adaptive, chen2023dbarf, yen2021inerf, smith2023flowcam, meuleman2023progressively, truong2023sparf, chng2022garf, park2023camp, hong2023unifying, kim2024up, xia2022sinerf} preforms the optimization on either an incremental or global manner.  
Incremental methods such as LocalRF~\cite{meuleman2023progressively} and NICER-SLAM~\cite{zhu2023nicer} achieve tracking of arbitrarily long sequences and handle large camera motions. However, these methods suffer from severe forgetting issues when using a single MLP~\cite{sucar2021imap}, or face difficulties in assigning multiple models~\cite{meuleman2023progressively, chen2023dbarf, wu2023scanerf}. 
In contrast, globally solving for poses in the NeRF optimization process can easily lead to local minima.
Earlier works have addressed this issue through coarse-to-fine positional encoding~\cite{lin2021barf}, sinusoidal activation for radiance mapping~\cite{xia2022sinerf}, or by restricting camera motions to forward-facing scenes~\cite{wang2021nerf}.
Generalizable unposed NeRF~\cite{song2023sc, Sajjadi2022RUSTLN, chen2023dbarf} methods enable the transferring of knowledge between different scenes even with no pose priors, however, they require learning much heavier networks such as a Vision Transformer~\cite{dosovitskiy2020image}. 
Recently, geometric priors such as monocular depth~\cite{bian2023nope}, optical flow~\cite{smith2023flowcam, smith2024flowmap} are widely used to regularize poses between adjacent frames. 
While these geometric priors can be effective in certain scenarios, they heavily depend on the quality of pre-trained geometric models, which can impose an upper-bound effect.
In contrast, we learn camera poses, scene geometry, and dense optical flow all on the fly, unifying all the geometry-related tasks under the same proposed NeRF scene representation framework.

\begin{figure*}[h]
  \centering \includegraphics[width=0.9\textwidth]{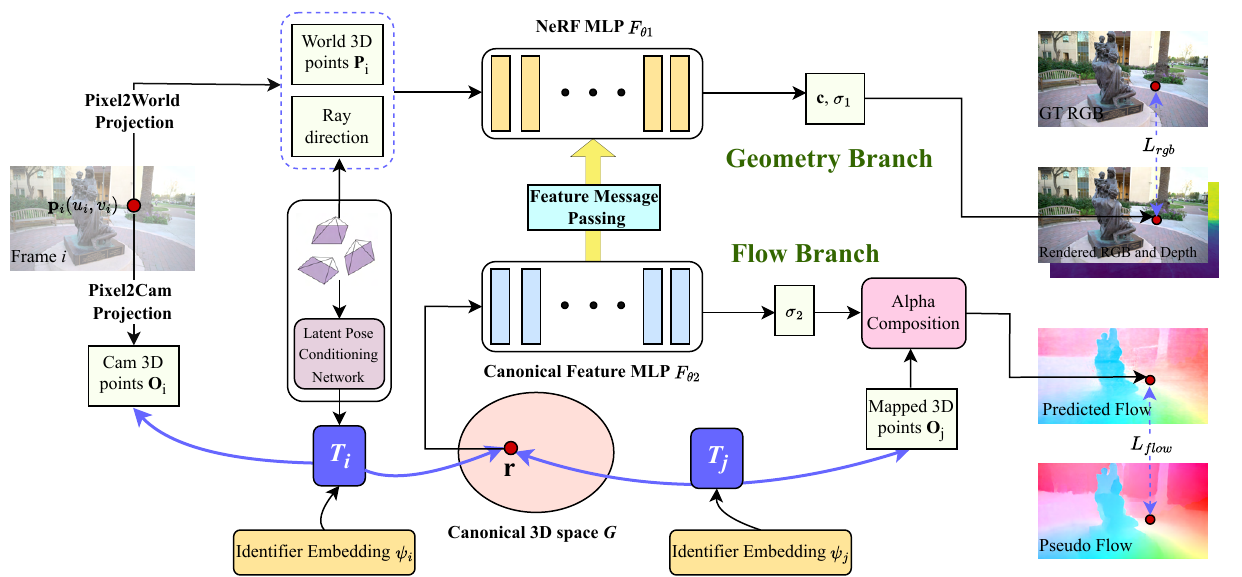}
  \vspace{-5pt}
  \caption{\textbf{Overview of the proposed Flow-NeRF}: Our method takes a sequence of images as input and jointly learns camera poses, scene geometry, and dense optical flow with a unified neural representation framework. We propose a shared points sampling mechanism to ensure the feature consistency between the geometry and flow branches~(Sec.~\ref{shared points}). We build a bijective mapping to query per-pixel motion given sampled points as input, conditioned on pose~(Sec.~\ref{query flow}). Leveraging the complementary nature of features between the world space and the 3D canonical volume, we enhance the feature representation of the geometry branch by message passing~(Sec.~\ref{feature enhancement}). We also develop effective loss functions to simultaneously learn flow and scene reconstruction, while imposing constraints on relative poses~(Sec.~\ref{loss function}). }
  \label{fig:2}
  \vspace{-10pt}
\end{figure*}

\par\noindent\textbf{Learning Dense Optical Flow.} Traditional methods typically solve dense optical flow estimation as an optimization problem. 
Learning-based flow estimation network such as FlowNet~\cite{dosovitskiy2015flownet}~\cite{hui2018liteflownet}, PWC-Net~\cite{sun2018pwc}, GMFlow~\cite{xu2022gmflow} and GMFlow++~\cite{xu2023unifying} have enhanced the quality and efficiency of the optical flow estimation. 
However, all of these methods are suitable for learning flow between consecutive frames, when querying flow between long-range pixels, their performance drops significantly as temporal inconsistency occurs. 
A benchmarking and leading method for flow estimation is RAFT~\cite{teed2020raft}, which learns flow through an iterative 4D correlation module. We use RAFT between adjacent frames as our pseudo-flow supervisory goal and outperforms RAFT on long-range dense flows with the proposed novel-view flow estimation in our framework.
Recently, a few methods~\cite{zhao2022particlesfm,harley2022particle,wang2023co,karaev2023cotracker,lai2021video,doersch2023tapir} emerged to solve the tracking problem in video sequence. For example, 
ParticleSfM~\cite{zhao2022particlesfm} optimizes long-range video correspondence as dense point trajectories.
Different from the above methods, Omnimotion~\cite{wang2023tracking} proposes a novel perspective that treats tracking of any points as the learning of a corresponding field based on neural representations. However, their method can only be used to query corresponding points on training views, which is not suitable for novel view synthesis. 
RUST~\cite{Sajjadi2022RUSTLN} introduces a pose-free approach to novel view synthesis, with the key insight of training a pose encoder that analyzes the target image to learn a latent pose embedding. This embedding can then be utilized by the decoder for effective view synthesis.
Inspired by Omnimotion and RUST, our work integrates, for the first time, a neural representation-based module that learns dense flow fields into a pose-free NeRF model. This integration allows for the joint learning of poses, scene geometry, and dense flow within a unified neural framework, enabling pose-conditioned novel-view flow estimation.

\section{The Proposed Flow-NeRF Method}

\begin{figure*}[!t]
\centering  \includegraphics[width=0.90\textwidth]{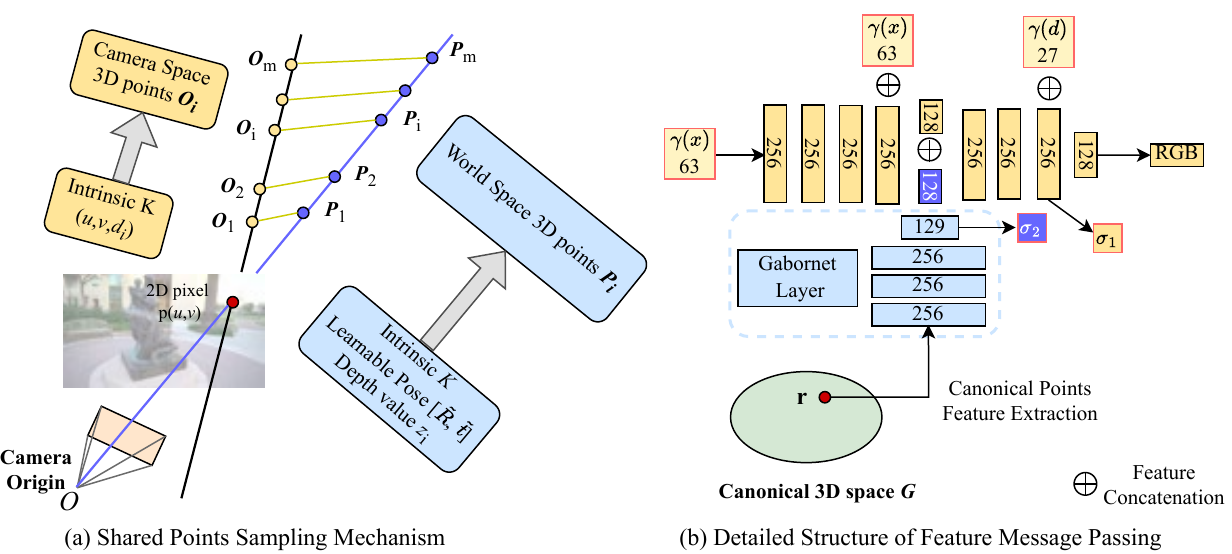}
\vspace{-5pt}
\caption{\textbf{Illustration of the details of the proposed shared points sampling mechanism} (a) for both the geometry and flow branches and the feature message passing module (b) to couple them for learning a unified scene neural representation.}
 \label{fig:3}
\vspace{-10pt}
\end{figure*}

Given a sequence of unposed images $\{I_1, I_2,..., I_{k-1}, I_{k}\}$, our goal is to simultaneously recover camera poses, scene geometry, and dense flow between consecutive or longer-range frames, all under a unified neural representation framework. Fig.~\ref{fig:2} overviews how the proposed Flow-NeRF framework works for the goal. It mainly consists of the geometry branch and the flow branch. During each iteration of training, we pick two frames $I_{i}$ and $I_{j}$, where $i<j$, and $\tau=j-i$ defines the interval between the selected two frames, which is set to 1 in most cases. 
We sample points on frame $i$ and use frame $j$ as the reference frame to query corresponding points and construct inter-frame losses. Specifically, we develop a shared points sampling strategy (see Fig.~\ref{fig:3}(a) and Sec.~\ref{shared points}) to ensure that every perspective 3D point in the world space for the geometry branch contains information of the same scene area encoded by the camera space 3D point for the flow branch.
Since we tackle static scenes, we assume that all the scene flow is induced by camera pose changes, and thus develop two latent space embeddings $\psi_{i}$ and $\psi_{j}$, which take poses of the two frames as input to condition on the flow prediction based on the bijective mapping $T_i$ and $T_j$ (see Sec.~\ref{query flow}). After mapping a camera space 3D point $\mathbf{O}_{i}$ to the canonical 3D space $G$, we obtain a 3D point $\mathbf{r}$ in the canonical space. Our key insight is that, although point $\mathbf{r}$ is not in the world space, it must share a similar underlying representation as it represents the same physical scene. Therefore, we develop a feature message passing strategy (see Sec.~\ref{feature enhancement} and Fig.~\ref{fig:3}(b)) to enhance the point feature in the world space from the extracted feature from the canonical point $\mathbf{r}$. The point features in the different spaces are learned based on different optimization losses and thus they are essentially complementary. 
More details about the losses are illustrated in Sec.~\ref{loss function}.         

\subsection{Preliminary: Pose-Free NeRF}
Neural Radiance Field (NeRF)~\cite{mildenhall2021nerf} represents a scene as an implicit mapping function $F_{\theta}: (\mathbf{x},\mathbf{d})\rightarrow(\mathbf{c}, \sigma)$ that maps a 3D point $\mathbf{x}$ and a viewing direction $\mathbf{d}$ to a radiance color $\mathbf{c}$ and a density value $\sigma$. The mapping function $F_{\theta}$ is parameterized as an MLP network with $\theta$ as the learnable parameters. Given a sequence of images $\{I_1, I_2,..., I_{k-1}, I_{k}\}$, pose-free NeRF learns the transformation from camera to world of every image $\{{T}_{1}, {T}_{2},...,{T}_{k-1}, {T}_{k}\}$ together with the MLP parameters $\theta$ to minimize the photometric loss between the rendered image and the ground truth image. The learnable transformation matrix (i.e, the camera pose) can be parameterized as a vector $\mathbf{v}=[r_1, r_2, r_3, t_1, t_2, t_3]$ of 6-degree of freedom using the axis-angle representation, and can be transformed to a pose matrix ${T}=[{R}, {t}]$ using Rodrigues' formula on the rotational part. 

Given the camera pose ${T}$ and a pixel coordinate $[u,v]$, NeRF back-projects a normalized viewing directing $\mathbf{d}$ and transforms it into world coordinate $\mathbf{r}=TK^{-1}[u,v]$, where $K$ is the known camera intrinsic matrix. Then it samples points along the ray with the corresponding depth values $\{z_1, z_2, ..., z_{m-1}, z_{m}\}$ between near and far bound $z_{n}$ and $z_{f}$, as $\mathbf{p}_{i}=z_{i}\mathbf{r}$. Both the color value $\hat{I}$ and the depth value $\hat{D}$ of $\mathbf{p}$ can then be rendered out with the volumetric rendering function:
\begin{equation}
\hat{I}(\mathbf{p})=\int_{z_{n}}^{z_{f}}T(z)\sigma(\mathbf{p}(z))\mathbf{c}(\mathbf{p}(z),\mathbf{d})dz
\end{equation}
\vspace{-10pt}
\begin{equation}
\hat{D}(\mathbf{p})=\int_{z_{n}}^{z_{f}}T(z)\sigma(\mathbf{p}(z))dz,
\end{equation}
where $T(z) = \text{exp}(-\int_{z_{n}}^{z}\sigma(\mathbf{p}(s))ds)$ is the accumulated transmittance along a ray.

\subsection{Shared Points Sampling for Geometry and Flow}\label{shared points}
To enable points sampled for the flow and the geometry branches in our framework to represent the same physical scene points, we design a shared points sampling mechanism, which ensures that we sample not only the same pixel points on the pixel plane but also the most relevant points in 3D for both branches. Specifically, we randomly sample $N$ 2D pixels, i.e., $\{p_1(u_1, v_1), p_2(u_2, v_2),..., p_N(u_N, v_N)\}$, during each training iteration on frame $i$. These sampled 2D pixels are shared for both branches, and then back-projected to the 3D space, using the pixel-to-world projection for the geometry branch, and the pixel-to-camera projection for the flow branch. In all our experiments, we sample $N=1024$ pixels (i.e., rays) for each frame.

Take a sampled 2D pixel coordinate $(u,v)$ as an example. As shown in Fig.~\ref{fig:3}(a), for the geometry branch, we follow the original NeRF to back-project the 2D pixel to the 3D world space using the known camera intrinsic matrix $K$, the learnable camera poses $T$, and several distance values $\{z_1, z_2, ..., z_{m-1}, z_{m}\}$ along the normalized ray direction $\mathbf{r}$, where $\mathbf{r}=TK^{-1}[u,v]$. Hence, the world space 3D points can be expressed as $\mathbf{p}_{i}=z_{i}\mathbf{r}$, where $z_{i}$ belongs to $\{z_1, z_2, ..., z_{m-1}, z_{m}\}$ that is between $z_{n}$ and $z_{f}$. 

For the flow branch, since only 2D-2D correspondences are learned, it is natural to follow the work of~\cite{wang2023tracking} to simply use an orthogonal projection to back-project the 2D pixels. However, this simplest orthogonal projection will lose the perspective relations of the scene, and thus significantly downgrade the depth estimation performance when doing feature message passing (see sec~\ref{feature enhancement}). We alternatively, propose to back-project the 2D pixels to camera space with the known intrinsic $K$. We experimentally find it can produce a much better depth map, while not sacrificing the accuracy of flow. Specifically, given the same set of 2D pixels $\{p_1(u_1, v_1), p_2(u_2, v_2),..., p_N(u_N, v_N)\}$ sampled for the geometry branch, we back-project a 2D pixel $(u,v)$ to the camera space, i.e, $K^{-1}[u,v]d_{i}$, where $d_{i}$ belongs to the set of depth values $\{d_1, d_2, ..., d_{m-1}, d_{m}\}$. Note that we expect the 3D points for both branches to represent the same physical scene points. To enable this, we fix the ratio of the distance value $z_{i}$ of the geometry branch and the depth value $d_{i}$ of the flow branch as follows:
\begin{equation}
    z_{i}=\alpha{d_{i}}, \ \ d_i \in \{d_1, d_2, ..., d_{m-1}, d_{m}\},
\end{equation}
where $\alpha$ is a constant. We set $\alpha=0.2$ in our experiments which means that all the input points for the flow branch are in the depth range of $[0.002,2]$.

\subsection{Learning Pixel-Wise Novel-View Dense Flow}\label{query flow}
Motivated by~\cite{wang2023tracking}, we formulate the prediction of 2D-2D correspondences between two frames as learning the bijective mapping $T_{i}$ and $T_{j}$. Specifically, $T_{i}$ maps the sampled 3D camera space point $\mathbf{O} _{i}$ of frame $i$ to a canonical 3D point $\mathbf{r}$, and $T_{j}$ maps $\mathbf{r}$ back to a 3D point $\mathbf{O}_{j}$ in the camera space of frame $j$ (see Fig.~\ref{fig:2}). We parameterize the bijective mapping as $\epsilon$ with an invertible neural network Real-NVP~\cite{dinh2016density}, as it theoretically guarantees reversibility. That is, for any given 3D point coordinate input $\mathbf{O}_{i}=K^{-1}[u,v]d$, after the bijective mapping, the output 3D point $\mathbf{O}_{j}$ has the same coordinates $K^{-1}[u,v]d$. The detailed invertible network structure is in the supplement.        

\begin{figure}[!t]
\centering \includegraphics[width=0.45\textwidth]{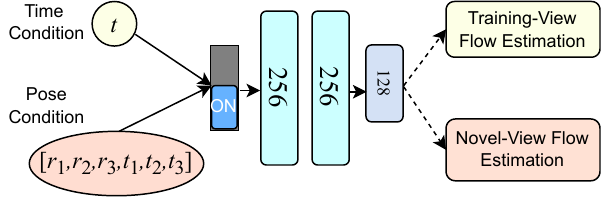}
\caption{\textbf{Latent feature embedding with time or pose condition}. While time-conditioned input can only infer on train views, our pose-conditioned input can infer novel-view flow thanks to the geometry certainty that pose can produce.}
\label{fig:4}
\vspace{-15pt}
\end{figure}

Given the bijective mapping, predicting 2D points displacement can be converted to learning the difference between $\mathbf{O}_{i}$ and $\mathbf{O}_{j}$. These differences are attributed to optical flow, which is caused by a mixture of camera and object movements. Omnimotion~\cite{wang2023tracking} models all these motions together, and proposes time as the unique identifier for discriminating different frames. We argue that although time-conditioned flow can learn a mixture of motions, it can be only applied to training views, because the time itself contains no physical information about the camera motion, and hence cannot be interpolated to infer flow on novel views. Given that our scenes are nearly static, we propose the pose-conditioned flow as we assume that the flow is induced by camera motion only. 
Fig.~\ref{fig:4} depicts the difference between time-conditioned and pose-conditioned inputs. Specifically, we build a network to embed the identifier of frame $i$ and frame $j$ to 128-dimensional features $\psi{i}$ and $\psi{j}$ in the latent space, respectively. For the pose-conditioned flow, we use the 6 DoF pose vector $[r_{1}, r_{2}, r_{3}, t_{1}, t_{2}, t_{3}]$ as a condition that depicts the `current' pose of the input frame. It changes along with the process of the pose optimization.  

After obtaining the series of pose-conditioned 3D points $\mathbf{O}_{j}(\mathbf{O}_{i}|\epsilon,\psi{i},\psi{j})$ of the same ray, we use alpha composition to compute the exact 3D point at frame $j$. The $\sigma$ value, denoted as $\sigma_{2}$ in Fig.~\ref{fig:2}, is obtained from the 3D canonical space point $\mathbf{r}$, given the canonical feature MLP $F_{\theta2}$ as the feature extractor. For the $k$-th sampled point along the ray, we compute the $\alpha$ value as $\alpha_{k}=1-\text{exp}(-\sigma_{k})$, and compute the predicted 3D point as follows:
\begin{equation}
\hat{\mathbf{O}}_{j}=\sum_{k=1}^{m}T_{k}\alpha_{k}\mathbf{O}_{j}^{k}, \ \ \ \ T_{k}=\prod_{l=1}^{k-1}(1-\alpha_{l}).
\end{equation}
The predicted 3D point $\hat{\mathbf{O}_{j}}$ is then projected to the 2D-pixel plane with camera intrinsic $K$ to obtain the predicted corresponding 2D pixel $\hat{\mathbf{p}}_{j}$ of frame $j$. 

\subsection{Unified Neural Scene Representation through Geometry and Flow Message Passing}\label{feature enhancement}

Intuitively, given the shared points sampling strategy in Sec.~\ref{shared points}, the point feature extracted from canonical space $G$ of the flow branch should share a lot in common with the point feature from the geometry branch, because both features represent the same physical scene. 
 
While these two branches utilize different loss functions for the optimization: the geometry branch is learned based on the photometric appearance rendering loss and the flow branch is learned based on pseudo motion flow supervision. These distinct supervision signals can make the two branches learn complementary features to represent the same physical scene. Besides, the provided 2D correspondences from the flow prediction can also bring benefits to more accurate geometry estimation to the geometry branch. A verification experiment regarding this complementarity is also shown in the supplement. 

Based on the above-discussed motivation, we thus design a simple yet effective message-passing strategy to propagate the features extracted from the canonical representation space to the world representation space parameterized by the MLP network $F_{\theta1}$ of the geometry branch. As shown in Fig.~\ref{fig:3}(b), we use a 3-layer of 256-dimensional embedding network, i.e., the Gabornet~\cite{alekseev2019gabornet}, to extract the canonical-space point feature. It outputs a 129-dimensional feature map. The 128-dimensional features are directly concatenated to one of the intermediate layers of the MLP network $F_{\theta1}$ of the geometry branch. The layer being concatenated in $F_{\theta1}$ is right after the skip connection of the positional embedding input, which maps the $256 + 63$ dimension feature to a 128 dimension feature that is directly concatenated with the 128 dimension feature produced from the canonical space point representation. We show from experiments that the message-passing mechanism effectively improves the scene geometry reconstruction by a large margin.   

\begin{figure*}[!t]
  \centering  \includegraphics[width=1\textwidth]{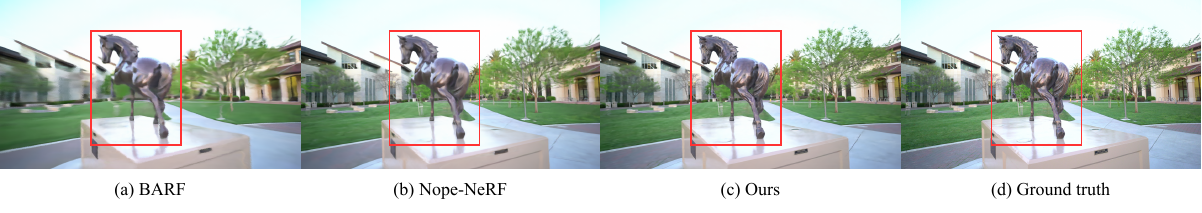}
  \vspace{-23pt}
  \caption{\textbf{Qualitative comparison with BARF~\cite{lin2021barf} and Nope-NeRF~\cite{bian2023nope} on novel view synthesis on the Tanks and Temples dataset}. Our method achieves superior novel-view rendering quality with enhanced details.}
  \label{fig:6}
  \vspace{-5pt}
\end{figure*}

\begin{table*}[!t]
\resizebox{\textwidth}{!}{
\begin{tabular}{ccccclccclccclccclccc}
\toprule[1pt]
\multirow{2}{*}{scenes} & & \multicolumn{3}{c}{Ours}                       &  & \multicolumn{3}{c}{Nope-NeRF~\cite{bian2023nope}} &  & \multicolumn{3}{c}{BARF~\cite{lin2021barf}} &  & \multicolumn{3}{c}{NeRFmm~\cite{wang2021nerf}} &  & \multicolumn{3}{c}{SC-NeRF~\cite{song2023sc}}              \\ \cline{3-5} \cline{7-9} \cline{11-13} \cline{15-17} \cline{19-21} 
                       & & PSNR$\uparrow$   & SSIM$\uparrow$          & LPIPS$\downarrow$         &  & PSNR     & SSIM    & LPIPS    &  & PSNR    & SSIM  & LPIPS  &  & PSNR    & SSIM   & LPIPS   &  & PSNR   & SSIM & LPIPS \\ \midrule[1pt]
\multicolumn{1}{c}{\multirow{5}{*}{\rotatebox[origin=c]{90}{ScanNet}}} & 0079\_00 &\textbf{34.02}    &\textbf{0.87}  &\textbf{0.34}    && 32.47    & 0.84    & 0.41    &&32.31     &0.83     &0.43     &&30.59     &0.81     &0.49     &&31.33     & 0.82    &0.46 \\
\multicolumn{1}{c}{} & 0418\_00 &\textbf{31.36}    &\textbf{0.79}  & 0.35    && 31.33    & \textbf{0.79}    & \textbf{0.34}    &&31.24     &\textbf{0.79}     &0.35     &&30.00     &0.77     &0.40     &&29.05     & 0.75    &0.43\\
\multicolumn{1}{c}{} & 0301\_00 &\textbf{30.22}    &\textbf{0.79}  & \textbf{0.34}    && 29.83    & 0.77    & 0.36    &&29.31     &0.76     &0.38    &&27.84     &0.72     &0.45     &&29.45     & 0.77    &0.35\\
\multicolumn{1}{c}{} & 0431\_00 &\textbf{34.60}    &\textbf{0.93}  & \textbf{0.32}    && 33.83    & 0.91    & 0.39    &&32.77     &0.90     &0.41    &&31.44     &0.88     &0.45     &&32.57     & 0.90    &0.40\\
\multicolumn{1}{c}{} & mean &\textbf{32.55}    &\textbf{0.85}  & \textbf{0.34}    && 31.86    & 0.83    & 0.38    &&31.41     &0.82     &0.39    &&29.97     &0.80     &0.45     &&30.60     & 0.81    &0.41\\ \midrule[1pt]
\multicolumn{1}{c}{\multirow{9}{*}{\rotatebox[origin=c]{90}{Tanks and Temple}}}
& Church                  & \textbf{28.27} & \textbf{0.83} & \textbf{0.28} &  & 25.17    & 0.73    & 0.39     &  & 23.17   & 0.62  & 0.52   &  & 21.64   & 0.58   & 0.54    &  & 21.96                     & 0.60 & 0.53  \\
\multicolumn{1}{c}{} & Barn                    & \textbf{28.53} & \textbf{0.78} & \textbf{0.35} &  & 26.35    & 0.69    & 0.44     &  & 25.28   & 0.64  & 0.48   &  & 23.21   & 0.61   & 0.53    &  & 23.26                     & 0.62 & 0.51  \\
\multicolumn{1}{c}{} & Museum                  & \textbf{29.43} & \textbf{0.85} & \textbf{0.27} &  & 26.77    & 0.76    & 0.35     &  & 23.58   & 0.61  & 0.55   &  & 22.37   & 0.61   & 0.53    &  & 24.94                     & 0.69 & 0.45  \\
\multicolumn{1}{c}{} & Family                  & \textbf{29.40} & \textbf{0.85} & \textbf{0.29} &  & 26.01    & 0.74    & 0.41     &  & 23.04   & 0.61  & 0.56   &  & 23.04   & 0.58   & 0.56    &  & 22.60                     & 0.63 & 0.51  \\
\multicolumn{1}{c}{} & Horse                   & \textbf{28.57} & \textbf{0.86} & \textbf{0.23} &  & 27.64    & 0.84    & 0.26     &  & 24.09   & 0.72  & 0.41   &  & 23.12   & 0.70   & 0.43    &  & 25.23                     & 0.76 & 0.37  \\
\multicolumn{1}{c}{} & Ballroom                & \textbf{28.83} & \textbf{0.86} & \textbf{0.24} &  & 25.33    & 0.72    & 0.38     &  & 20.66   & 0.50  & 0.60   &  & 20.03   & 0.48   & 0.57    &  & 22.64                     & 0.61 & 0.48  \\
\multicolumn{1}{c}{} & Francis                 & \textbf{30.63} & \textbf{0.83} & \textbf{0.33} &  & 29.48    & 0.80    & 0.38     &  & 25.85   & 0.69  & 0.57   &  & 25.40   & 0.69   & 0.52    &  & 26.46                     & 0.73 & 0.49  \\
\multicolumn{1}{c}{} & Ignatius                & \textbf{26.25} & \textbf{0.73} & \textbf{0.35} &  & 23.96    & 0.61    & 0.47     &  & 21.78   & 0.47  & 0.60   &  & 21.16   & 0.45   & 0.60    &  & 23.00                     & 0.55 & 0.53  \\
\multicolumn{1}{c}{} & mean                     & \textbf{28.73} & \textbf{0.82} & \textbf{0.29}               &  & 26.34    & 0.74    & 0.39     &  & 23.42   & 0.61  & 0.54   &  & 22.50   & 0.59   & 0.54    &  & \multicolumn{1}{l}{23.76} & 0.65 & 0.48  \\ \bottomrule[1pt]
\end{tabular}
}
\vspace{-6pt}
\caption{\textbf{Quantitative comparison results of Novel view synthesis on the Tanks and Temples and ScanNet datasets}. Our method outperforms other state-of-the-art methods by a large margin.}
\label{table:1}
\vspace{-15pt}
\end{table*}

\subsection{Optimization Loss Functions of Flow-NeRF}\label{loss function}

The geometry branch renders RGB images, and we utilize the photometric consistency loss function. 

For the flow branch, we apply a flow loss between the predicted 2D pixel $\hat{\mathbf{p}}_{j}$ and the ground truth 2D pixel $\mathbf{p}_{j}$ provided by the pseudo flow computed from RAFT~\cite{teed2020raft}. These two loss functions are detailed as follows: 
\begin{equation}
L_{rgb}=\frac{1}{N}\sum_{\mathbf{p}\in{\Omega_{N}}}{||\mathbf{\hat{C}}(\mathbf{p})-\mathbf{C}(\mathbf{p})||_{1}}
\end{equation}
\vspace{-8pt}
\begin{equation}
L_{flow}=\frac{1}{N}\sum_{\mathbf{p}_j\in{\Omega_{N}}}||\hat{\mathbf{p}}_{j}-\mathbf{p}_{j}||_{1},
    \label{eq:7}
\end{equation}
where $\mathbf{\hat{C}}$ and $\mathbf{C}$ denote the rendered and the ground truth RGB image, respectively. We also consider 3D point cloud and 2D photometric warping losses. Please refer to the supplement for details. The geometry and flow branches of Flow-NeRF are jointly optimized with these losses.

\section{Experiments}

\begin{figure*}[ht]
  \centering
\includegraphics[width=1.0\textwidth]{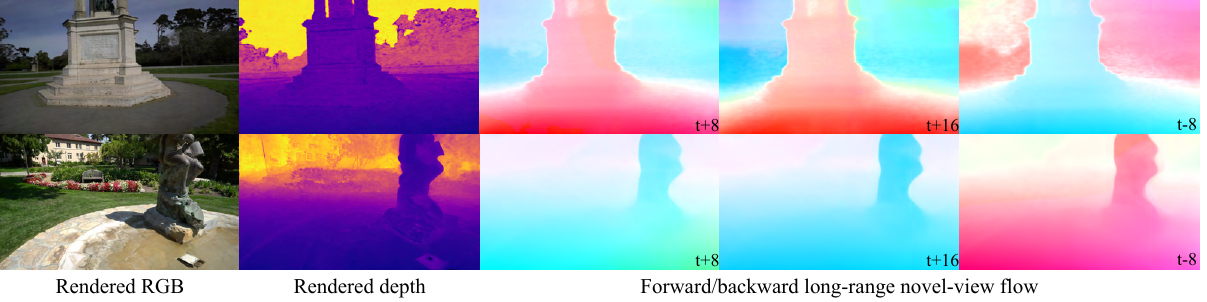}
  \vspace{-20pt}
  \caption{\textbf{Novel-view flow estimation on long-range frames on the Tanks and Temples dataset}. We utilize only consecutive forward flow for training supervision; however, our model can infer plausible long-range flow in both forward and backward directions. t+8 and t-8 denote forward and backward flow with a time interval of 8, respectively, while t+16 indicates forward flow with a time interval of 16.}
  \label{fig:novel1}
  \vspace{-6pt}
\end{figure*}

\begin{table*}[!t]
\resizebox{\textwidth}{!}{
\begin{tabular}{ccccclccclccclccclccc}
\toprule[1pt]
\multirow{2}{*}{scenes}  &  & \multicolumn{3}{c}{Ours}     &  & \multicolumn{3}{c}{Nope-NeRF~\cite{bian2023nope}}     &  & \multicolumn{3}{c}{BARF~\cite{lin2021barf}}     &  & \multicolumn{3}{c}{NeRFmm~\cite{wang2021nerf}}     &  & \multicolumn{3}{c}{SC-NeRF~\cite{song2023sc}}           \\ \cline{3-5} \cline{7-9} \cline{11-13} \cline{15-17} \cline{19-21} 
    & & $\text{RPE}_t\downarrow$    & $\text{RPE}_r\downarrow$    & ATE$\downarrow$    &  & $\text{RPE}_t$      & $\text{RPE}_r$      & ATE     &  & $\text{RPE}_t$    & $\text{RPE}_r$    & ATE    &  & $\text{RPE}_t$     & $\text{RPE}_r$     & ATE    &  & $\text{RPE}_t$      & $\text{RPE}_r$   & ATE   \\ \midrule[1pt]
\multicolumn{1}{c}{\multirow{5}{*}{\rotatebox[origin=c]{90}{ScanNet}}} & 0079\_00  & 0.251 & 0.348 & 0.010 & & 0.752 & 0.204 & 0.023 && 1.110 & 0.480 & 0.062 & & 1.706 & 0.636 & 0.100 && 2.064 & 0.664 & 0.115 \\
\multicolumn{1}{c}{} & 0418\_00  & 0.527  & 0.155 & 0.026 && 0.455  & 0.119  & 0.015  && 1.398 & 0.538 & 0.020      &&1.402       &0.460      & 0.013     && 1.528     &0.502      &0.016      \\
\multicolumn{1}{c}{} & 0301\_00  &0.186     &0.197      &0.007      &&0.399       &0.123      &0.013     &&1.316       & 0.777     &0.219      &&3.097       &0.894      &0.288      &&1.133      &0.422      &0.056      \\
\multicolumn{1}{c}{} & 0431\_00  &0.269     &0.355      &0.013      &&1.625       &0.274      &0.069     &&6.024       & 0.754     &0.168      &&6.799       &0.624      &0.496      &&4.110      &0.499      &0.205      \\
\multicolumn{1}{c}{} & mean  &\textbf{0.308}     &0.263      &\textbf{0.014}      &&0.808       &\textbf{0.180}      &0.030     &&2.462       &0.637      &0.117      &&3.251       &0.654      &0.224      & & 2.209      &0.522      &0.098      \\   \bottomrule[1pt]
\end{tabular}
}
\vspace{-8pt}
\caption{\textbf{Quantitative comparison results of camera poses on the ScanNet dataset}.  
Note that we evaluate against the ground truth poses of ScanNet, which provides a more reliable reference than using poses obtained from COLMAP preprocessing.}
\label{table:ScanNet-pose}
\vspace{-10pt}
\end{table*}

\subsection{Experimental Setup}
\textbf{Datasets:} We conduct experiments on three datasets: \emph{Tanks and Temples}~\cite{knapitsch2017tanks}, \emph{ScanNet}~\cite{dai2017scannet}, and \emph{Sintel}~\cite{Geiger2012CVPR}. Please see the appendix for more details on the benchmarks.

\noindent
\textbf{Metrics:} For \emph{novel view synthesis}, we evaluate on standard metrics, i.e., PSNR, SSIM~\cite{wang2004image}, and LPIPS~\cite{zhang2018unreasonable}. For~\emph{camera pose}, we evaluate the absolute tracking error (ATE) on the sequence of all estimated poses, the relative rotation error (RPE$_{r}$) and translation error (RPE$_{t}$) between consecutive image pairs. The estimated camera poses are aligned with the ground truth poses using Umeyama alignment~\cite{umeyama1991least} before evaluation. For \emph{novel-view depth}, we use standard depth metrics (i.e., Abs Rel, Sq Rel, RMSE, RMSE log, $\delta_{1}$, $\delta_{2}$, and $\delta_{3}$). We follow~\cite{zhou2017unsupervised} to recover the metric depth values based on matching the median values between the predicted and ground truth depth. For \emph{novel-view flow}, we use average end-point error (EPE) to evaluate the predicted flow against the ground truth flow. 
Details of the evaluation protocols can be found in the appendix. 

\subsection{State-of-the-art Comparison}
We compare our approach with previous state-of-the-art pose-unknown methods on the sub-tasks we modeled. 

\noindent
\textbf{Novel-view synthesis:} Following the practice of Nope-NeRF~\cite{bian2023nope}, 
we initialize the novel-view poses from adjacent training-view poses and then optimize these poses by minimizing the photometric loss between the rendered image and the test-view image. 
The quantitative and qualitative comparison results are presented in Tab.~\ref{table:1} and Fig.~\ref{fig:6}. Our method outperforms all other state-of-the-art methods on both the ScanNet and Tanks and Temples datasets. The qualitative results demonstrate that our method can render more photo-realistic novel scenes while preserving more fine-grained details in the images. 

\noindent
\textbf{Depth prediction:} Tab.~\ref{table:depth} summarizes the quantitative results for novel-view depth on the ScanNet dataset. Our method demonstrates significant performance improvements across all depth metrics compared to other competitive methods. The qualitative depth comparisons and per-scene results are available in the supplementary material.

\noindent
\textbf{Pose estimation:} Our method demonstrates significantly better pose prediction than BARF~\cite{lin2021barf}, NeRFmm~\cite{wang2021nerf}, and SC-NeRF~\cite{song2023sc} on the ScanNet dataset (see Tab.~\ref{table:ScanNet-pose}), while performing on par with the state-of-the-art method Nope-NeRF~\cite{bian2023nope}.
Importantly, our method excels Nope-NeRF in terms of the metrics ATE and RPE$_{t}$.

\begin{table}[!t]
 \centering
\setlength{\tabcolsep}{2pt}
\resizebox{\linewidth}{!}{
\begin{tabular}{lccccccc}
\toprule[1pt]
Method       & Abs Rel $\downarrow$ & Sq Rel $\downarrow$ & RMSE $\downarrow$ & RMSE log $\downarrow$ & $\delta_1 \uparrow$ & $\delta_2 \uparrow$ & $\delta_3 \uparrow$ \\ \midrule[1pt]
BARF~\cite{lin2021barf}  &0.376  &   0.684  &   0.990  &   0.401  &   0.490  &   0.751  &   0.884                        \\
NeRFmm~\cite{wang2021nerf} & 0.590  &   1,721  &   1.672  &   0.587  &   0.316  &   0.560  &   0.743 \\
SC-NeRF~\cite{song2023sc} & 0.417  &   0.642  &  1.079  &   0.476  &   0.362  &   0.658  &   0.832
\\
Nope-NeRF~\cite{bian2023nope}    & 0.141  &   0.137  &   0.568  &   0.176  &   0.828  &   0.970  &   0.987                     \\
Ours    & \textbf{0.047}       & \textbf{0.013}      & \textbf{0.151}    & \textbf{0.072}        & \textbf{0.982}                   & \textbf{0.993}                     & \textbf{0.999}                      \\ \bottomrule[1pt]
\end{tabular}
}
\vspace{-8pt}
\caption{\textbf{Quantitative comparison results of depth estimation on the ScanNet dataset}. Our method achieves significantly higher depth accuracy compared to all other evaluated approaches.}
\label{table:depth}
\vspace{-10pt}
\end{table}

\subsection{Model Analysis}
\textbf{Flow-enhanced novel-view synthesis}
Our implicit unified representation significantly enhances the novel-view synthesis quality of the geometry branch, as demonstrated in Tab.~\ref{table:4}.
On the Ballroom scene, the PSNR metric increases by approximately 4 points. 
This is because the flow branch shares the same underlying yet complementary feature representation with the geometry branch, and thus, through the proposed feature message passing, the geometry branch can learn more structural details of the scene. This is further evidenced in the supplementary document.

\begin{figure}[!t]
    \centering   \includegraphics[width=\linewidth]{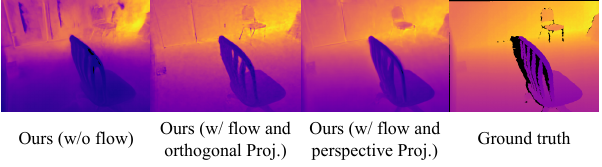}
    \vspace{-20pt}
    \caption{\textbf{Geometry enhancement through the flow branch on the 0418$\_$00 scene of ScanNet.} Our method predicts smoother depth maps with significantly fewer artifacts compared to variants that do not incorporate flow message passing.
    In contrast to the orthogonal projection sampling in Omnimotion~\cite{wang2023tracking}, our perspective projection effectively preserves correct 2D-3D relationships, resulting in improved depth map quality.}
    \label{fig:depth}
    \vspace{-10pt}
\end{figure}

\noindent
\textbf{Flow-enhanced geometry} 
Previous methods~\cite{lao2023corresnerf, smith2023flowcam} that utilize flow-assisted NeRF explicitly regularize the relationships between camera pose and depth through correspondence priors.
Although effective, these approaches do not fully leverage the advantages that a neural field can provide.
We argue that the flow map not only offers explicit correspondence cues but also contains rich texture information that can be implicitly integrated into a unified field representation.
Specifically, due to illumination variance in the images and inaccuracies in pose estimation during optimization, depth maps predicted without flow enhancement exhibit several visible artifacts, as shown in Fig.~\ref{fig:depth}.
In contrast, our flow enhancement design effectively improves the local smoothness of the produced depth maps through feature message passing, significantly reducing artifacts.
Furthermore, our innovative design of the perspective projection for point sampling in the flow branch helps preserve the correct 2D-3D relationships within the geometry branch, thus leading to better depth map accuracy, as shown in Fig.~\ref{fig:depth} and Tab.~\ref{table:depth_compare}.

\begin{table}[!t]
\centering
\setlength{\tabcolsep}{2pt}
\resizebox{0.8\linewidth}{!}{
\begin{tabular}{lccc}
\toprule[1pt]
Method                      & PSNR$\uparrow$  & SSIM$\uparrow$ & LPIPS$\downarrow$         \\ \midrule[1pt]
Nope-NeRF~\cite{bian2023nope}                & 25.33 & 0.72 & 0.38          \\
Ours w/o Flow Enhancement & 24.85 & 0.71 & 0.38          \\
Ours w/ Flow Enhancement  & \textbf{28.83} & \textbf{0.86} & \textbf{0.24}\\ \bottomrule[1pt]
\end{tabular}
}
\vspace{-8pt}
\caption{\textbf{Flow-enhanced novel-view synthesis on the Ballroom scene}. The feature distillation from the flow branch largely boosts the performance of novel-view synthesis.}
\label{table:4}
\vspace{-10pt}
\end{table}

\begin{figure*}[!t]
\centering
\includegraphics[width=0.95\linewidth]{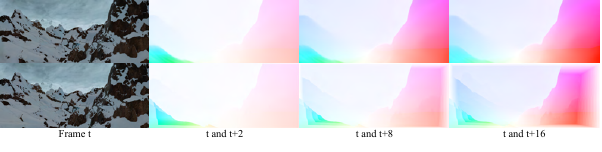}
\vspace{-15pt}
  \caption{\textbf{Novel-view flow comparison at different frame intervals against the ground-truth flow on the Sintel dataset}. As the frame interval increases, the rendered novel-view flow (the first row) appears in deeper shades of red, which is consistent with the behavior of the ground truth flow (the second row). These results suggest that our method can effectively infer plausible long-range novel-view flow.}
  \label{fig:sintel-flow}
  \vspace{-12pt}
\end{figure*}

\begin{table}[!t]
\setlength{\tabcolsep}{2pt}
\resizebox{\linewidth}{!}{
\begin{tabular}{lcccc}
\toprule[1pt]
Method       & Abs Rel $\downarrow$ & Sq Rel $\downarrow$ & RMSE $\downarrow$ & RMSE log $\downarrow$  \\ \midrule[1pt]
Ours w/o Flow Enhancement    & 0.200       & 0.206      & 0.809    & 0.221      \\
Ours w/ Flow Enhancement (Orthogonal)  & 0.047       & 0.011      & 0.147    & 0.075      \\
Ours w/ Flow Enhancement (Perspective)   & \textbf{0.034}  &   \textbf{0.010}  &   \textbf{0.118}  &   \textbf{0.070}  \\ \bottomrule[1pt]
\end{tabular}
}
\vspace{-8pt}
\caption{\textbf{Geometry enhancement through the flow branch on the 0418$\_$00 scene of ScanNet}. Our method with flow enhancement and perspective projection produces the best depth accuracy.}
\label{table:depth_compare}
\vspace{-15pt}
\end{table}

\noindent
\textbf{Holistic scene modeling through pose-conditioned novel-view flow}
The test-time optimization methods, such as omnimotion~\cite{wang2023tracking}
struggle to interpolate novel-view flow because they condition the model input on time, which introduces geometry ambiguities.
In contrast, our pose-conditioned design for the flow branch can be extended to \emph{novel views} under the unified optimization framework of both the geometry branch and flow branches. 
We define novel-view flow analogously to novel-view images in the context of NeRF.
During the inference, given novel poses from two arbitrary views as conditions for the model, it can render the flow for these two views. We provide several visualizations of the novel-view flow predictions for both forward and backward long-range frames (see Fig.~\ref{fig:novel1}). 
These results demonstrate that our method can produce qualitatively sound novel view flows. 
In our visualization of the long-range backward flow, although these frames are not used for supervision during training, we can still render plausible novel-view flow.
This capability is attributed to the pose-conditioned input design that successfully embeds all the pose-induced 2D displacements into the unified field representation. 
More visualization results on novel-view flow predictions can be found in the supplement.

\begin{figure}[!t]
\centering
\includegraphics[width=0.95\linewidth]{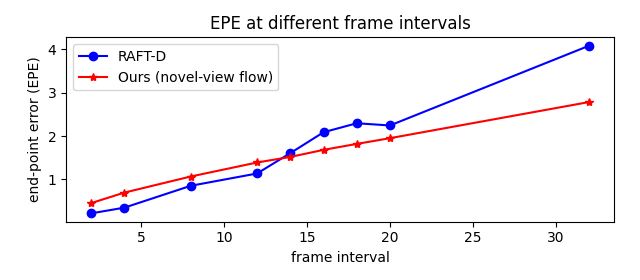}
  \vspace{-15pt}
  \caption{\textbf{Average end-point error (EPE) comparison on different frame intervals on the scene mountain1 of Sintel:} As the frame interval increases to 16, our method demonstrates the ability to infer more accurate flow than RAFT.}
  \label{fig:epe-time}
  \vspace{-15pt}
\end{figure}

\noindent
\textbf{Accurate long-range novel-view flow inference}
To quantitatively evaluate the quality of the predicted novel-view flows, we perform evaluations against the ground truth flow on the Sintel dataset. We compare our novel-view flow and RAFT flow prediction on average end-point error (EPE) at non-occluded pixels at different time intervals, as shown in Fig.~\ref{fig:epe-time}. While RAFT performs well between closer frames, its performance degrades a lot when the frame interval increases. Our method can infer more accurate longer-range flows on novel views. It should be noted that, although our method relies on RAFT for supervision, the RAFT prior is \emph{not an upper bound} of our method. This results suggest that learning dense flow within a unified neural field can indeed enhance long-range flow performance.
We also provide qualitative results of the predicted novel-view flows in Fig.~\ref{fig:sintel-flow}. We can observe that the predicted novel-view flows contain sharp details. As the frame interval increases, the flow values become larger, indicated by deeper red regions in the visualized flow image, suggesting that we have successfully predicted plausible longer-distance flows.
\vspace{-4pt}

\section{Conclusion}
We have presented Flow-NeRF, a unified neural representation framework to simultaneously learn camera poses, scene geometry, and dense optical flow. 
We propose two branches for the framework, incorporating a shared point sampling strategy that ensures sampled points accurately represent the same physical scene point.
As a result, we can pass feature messages from the flow branch to the geometry branch to enhance geometry feature learning. Our novel design on pose-conditioned flow enables quantitatively accurate long-range flow estimation and qualitatively sound flow rendering from novel views. Experiments on various datasets have shown that our method significantly improves the scene geometry, and yields plausible novel-view flow predictions.

{
    \small
    \bibliographystyle{ieeenat_fullname}
    \bibliography{main}
}
\clearpage
\clearpage
\setcounter{page}{1}
\maketitlesupplementary

\section{More details of the optimization losses}
Following Nope-NeRF~\cite{bian2023nope}, we enforce a depth loss between the rendered depth $\hat{D}(\mathbf{p})$ and the undistorted pre-computed pseudo ground truth depth $D^{\ast}(\mathbf{p})$ as follows:
\begin{equation}
    L_{depth}=\frac{1}{N}\sum_{\mathbf{p}\in{\Omega_{N}}}||D^{\ast}(\mathbf{p})-\hat{D}(\mathbf{p})||_{1}
\end{equation}
Additionally, we consider a point cloud loss to constrain the relative poses between frame $i$ and frame $j$: 
\begin{equation}
    L_{pc}=\sum_{(i,j)}l_{cd}(P^{\ast}_{j}, T_{ji}P^{\ast}_{i}),
\end{equation}
where $P^{\ast}_{j}$ and $P^{\ast}_{i}$ denote the point clouds of frame $j$ and frame $i$ computed from their undistorted depths $D^{\ast}_{j}$ and $D^{\ast}_{i}$, respectively; $T_{ji}$ represents the relative poses of the two frames; and $l_{cd}$ denotes the Chamfer Distance between the two point clouds.
Additionally, we introduce a photometric warping loss for the entire image, given by the projection of point cloud $P^{\ast}_{i}$ onto frame $j$:
\begin{equation}
L_{rgb\_s}=\sum_{(i,j)}||I_{i}\langle{K}_{i}P^{\ast}_{i}\rangle-I_{j}\langle{K}_{j}T_{j}T_{i}^{-1}P^{\ast}_{i}\rangle||_{1},
\end{equation}
where $\langle\rangle$ represents the bilinear interpolation operation on the image to acquire the corresponding color. Finally, our loss function is defined as:
\begin{equation}
L_{o} = L_{rgb}+\lambda_{1}L_{flow}+\lambda_{2}L_{depth}+\lambda_{3}L_{pc}+\lambda_{4}L_{rgb\_s}.
\end{equation}
The geometry and the flow branches of Flow-NeRF are jointly optimized using the overall loss $L_{o}$.

\section{Verification of feature complementarity between the geometry and the flow branches}
\label{complementary}
\begin{figure*}[h]
  \centering  
  \vspace{-10pt}\includegraphics[width=1.0\textwidth]{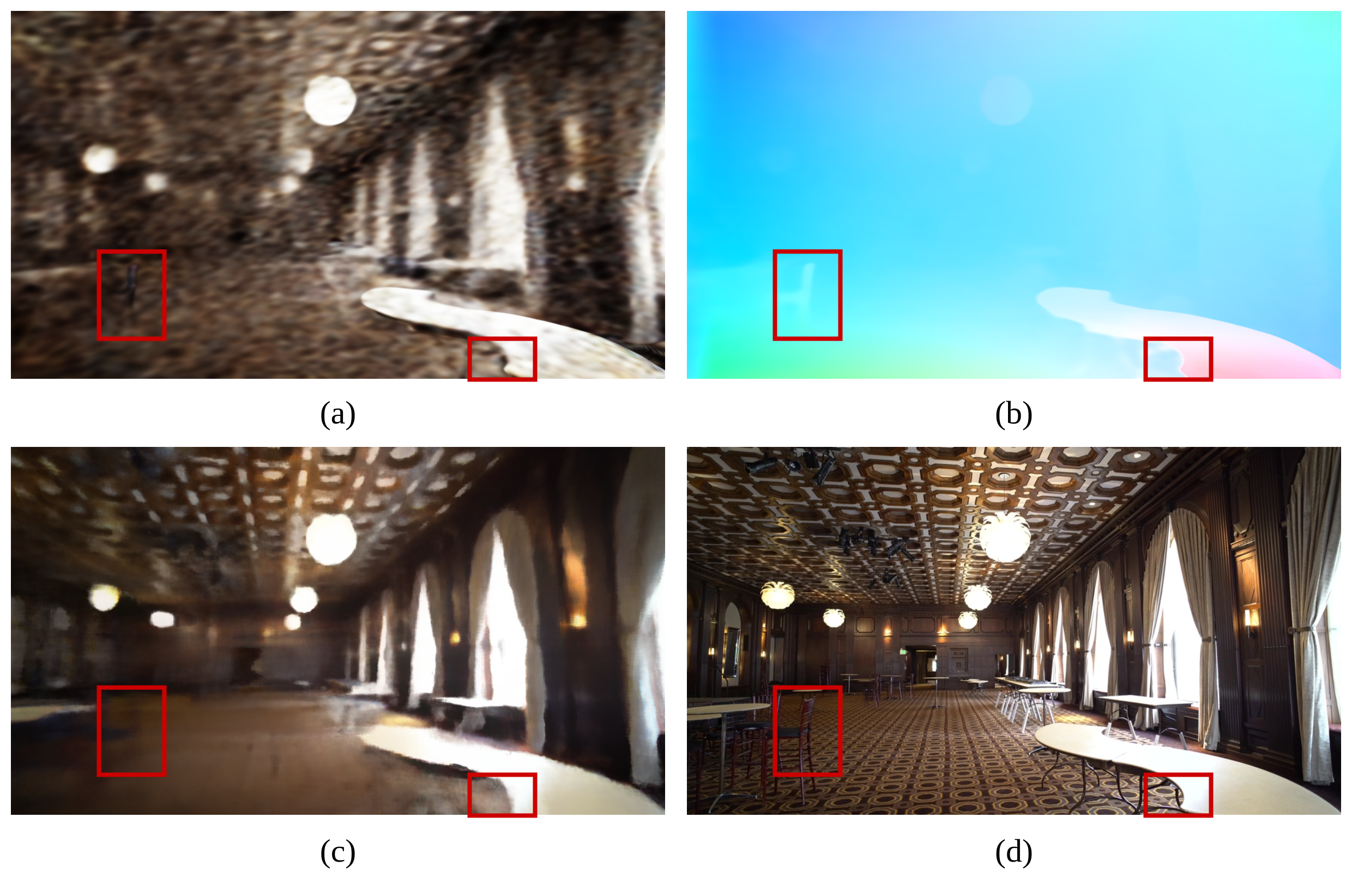}
  \vspace{-30pt}
  \caption{\textbf{Complementary features of geometry and flow branch outputs at training iteration 20000:} (a) RGB image rendered from the canonical feature $F_{\theta2}$; (b) flow prediction; (c) RGB image rendered from NeRF MLP $F_{\theta1}$; (d) RGB ground truth.}
  \label{fig:5}
  \vspace{-10pt}
\end{figure*}

Intuitively, given the shared points sampling strategy described in Sec.~3.2 of the main paper, the point features extracted from the canonical space $G$ of the flow branch should have significant overlap with the point features from the geometry branch, as both features represent the same physical scene. 
We validate this insight by rendering RGB images from the canonical feature extractor $F_{\theta2}$, resulting in a 4-channel tensor output. The first three channels correspond to RGB values, while the last channel predicts the alpha value $\sigma_{2}$. To visualize what can be learned from the canonical features, we enforce an additional photometric loss between the rendered flow RGB $\mathbf{\hat{C}}_{flow}$ and the ground truth RGB image $\mathbf{C}$ as:
\begin{equation}
    L_{rgb\_flow}=\frac{1}{N}\sum_{\mathbf{p}\in{\Omega_{N}}}{||\mathbf{\hat{C}}_{flow}(\mathbf{p})-\mathbf{C}(\mathbf{p})||_{1}},
\end{equation}
where $\Omega$ is the set of all $N$ pixels sampled from the frame. For the RGB prediction from the geometry branch, the loss function is defined as:
\begin{equation}
    L_{rgb}=\frac{1}{N}\sum_{\mathbf{p}\in{\Omega_{N}}}{||\mathbf{\hat{C}}(\mathbf{p})-\mathbf{C}(\mathbf{p})||_{1}},
    \label{eq:6}
\end{equation}
In Fig.~\ref{fig:5}, we visualize the results at iteration 20000, an early stage of the training process. It can be observed that within the red box, the RGB rendered from the flow branch and the flow map prediction show clearer details of the desk corner and the chair contour. In contrast, while exhibiting more visually realistic colors, the RGB image of the geometry branch lacks the accuracy of the geometry depicted in the flow RGB. This simple verification experiment demonstrates that the flow branch captures more structured and finer details than the geometry branch. This observation inspires us to develop the flow feature message passing strategy for the geometry branch, which proves effective in improving both the novel-view synthesis and depth prediction.

\section{Dataset details}
For \textbf{Tanks and Temples}, following Nope-NeRF~\cite{bian2023nope}, 
we evaluate novel view synthesis across 8 scenes. We sample 7 images from each 8-frame clip as training views and evaluate the novel view synthesis results on all other views. For the Family scene, we follow the work of Nope-NeRF~\cite{bian2023nope} by sampling every other view and evaluating the novel view synthesis results on the remaining half. 

For \textbf{ScanNet}, following Nope-NeRF~\cite{bian2023nope}, we sample 80-100 images from 4 scenes. We sample 7 training views from each 8-frame clip and evaluate both the novel view synthesis results and the depth estimation results. For data preprocessing, we use the ImageMagick~\cite{imagemagick} toolbox to downsample all images into half resolution. In addition, for Scene 0079$\_$00, we crop the dark borders by 10 pixels before preprocessing. The details of the selected ScanNet sequences are shown in Tab.~\ref{table: data}.

\begin{table}[ht]
\resizebox{\linewidth}{!}{
\begin{tabular}{lccccc}
\toprule[1pt]
                  & Scenes   & Type    & Seq. length & Frame ID & Max. rotation (deg) \\ \toprule[1pt]
\multirow{4}{*}{\rotatebox[origin=c]{90}{ScanNet}} & 0079\_00 & indoor  & 90           & 331-420            & 54.4     \\
                  & 0418\_00 & indoor  & 80           & 2671-2750            & 27.5     \\
                  & 0301\_00 & indoor  & 100          & 831-930            & 43.7     \\
                  & 0431\_00 & indoor  & 100          & 591-690            & 45.8     \\ \toprule[1pt]

\end{tabular}
}
\vspace{-10pt}
\caption{\textbf{Details of the selected ScanNet sequences.}}
\label{table: data}
\vspace{-15pt}
\end{table}

For \textbf{Sintel}, which consists of several scenes with ground truth flow available between consecutive frames, we train on 2 scenes: mountain1 and sleeping2. Each scene contains a total of 50 images; we sample every other frame to create the training set, leaving the other 25 frames as novel view test frames. For the flow evaluation,  we compare our novel-view flow results with the RAFT flow prediction results using the average end-point error (EPE) at non-occluded pixels across different frame intervals, as shown in Fig.~9 of the main paper. RAFT-D in Fig.~9 refers to directly inferring distant frame flows with RAFT, which yields better flow results than RAFT-C (where consecutive RAFT flow predictions are chained to formulate long-range flow). We obtain the ground truth optical flow for long-range flows by chaining consecutive ground truth flow along with their occlusion masks. We calculate the end-point error (epe) as follows:
\begin{equation}
\text{epe}=\frac{1}{M}\sum_{\mathbf{p}_{gt}\in{\Omega_{M}}}||\hat{\mathbf{p}}_{est}-\mathbf{p}_{gt}||_{1},
\vspace{-5pt}
\end{equation}
where $\hat{\mathbf{p}}_{est}$ and $\mathbf{p}_{gt}$ denote the estimated and the ground truth flow vector, respectively, and $\Omega_{M}$ is the set of all non-occluded pixels.

\section{Implementation details}
\textbf{Network structure and learning rate:}  For the bijective network, following omnimotion~\cite{wang2023tracking}, we use a simplified Real-NVP~\cite{dinh2016density} but with much fewer layers for the trade-off of training speed. We use 4 affine coupling layers for the network, and each layer contains three 128-dimensional MLPs. We set the initial learning rate of both the pose and the $F_{\theta2}$ MLP to be 0.0005, the bijective network to be 0.0001, the canonical feature MLP to be 0.0003 and the latent space embedding network to be 0.001. We employ an auto-scheduler to decrease the learning rate for both the network and the pose until the training PSNR does not increase for more than 1000 epochs. We set $\lambda_{1}=0.05$, $\lambda_{2}=0.04$, $\lambda_{3}=1$ and $\lambda_{4}=1$ for all the loss terms.

\noindent
\textbf{Point sampling:} For simplicity, we discard the hierarchical sampling strategy in the original NeRF, but apply a uniform sampling with perturbation during training to sample $m$ distance values. In all our experiments, we set the near and far bound of $z_{n}=0.01$, $z_{f}=10$, and $m=128$. 

\noindent
\textbf{Pose module:} The pose-conditioned latent embedding network is a 256-dimensional 3 layer MLP with Gaber layer, which takes a 6DOF pose vector [$r_1, r_2, r_3, t_1, t_2, t_3$] as input and outputs a 128-dimensional feature. The latent pose feature in the flow branch serves as an identifier of different frames. The driving force for pose learning comes from the complementary of RGB, flow, depth, and point cloud matching loss.

\begin{table}[!t]
\centering
\setlength{\tabcolsep}{2pt}
\resizebox{1.0\linewidth}{!}{
\begin{tabular}{lcccccccc}
\toprule[1pt]
     Method   &PSNR $\uparrow$ &SSIM $\uparrow$ & Abs Rel $\downarrow$ & RMSE $\downarrow$ & $\delta_1$ $\uparrow$ & $\delta_2$ $\uparrow$   & $\delta_3$ $\uparrow$  \\ \midrule[1pt]
LocalRF (Meuleman, 2023)    &31.25  &0.83 &11.148  &1.498 &0.422  &0.564  &0.850                        \\
CF-3DGS (Yu, 2023)    &28.51  &0.80 &12.360 &1.014  &0.617  &0.819  &0.875
          \\
Ours    & \textbf{32.55}  &\textbf{0.85}  &\textbf{0.047} &\textbf{0.151} &\textbf{0.982} &\textbf{0.993} &\textbf{0.999}            \\ \bottomrule[1pt]
\end{tabular}
}
\vspace{-8pt}
\caption{\textbf{Comparison against SOTA methods on novel-view synthesis and depth estimation across all 4 scenes on the ScanNet dataset.}}
\label{table:depth_compare}
\vspace{-5pt}
\end{table}

\begin{table}[!t]
\vspace{-2pt}
\centering
\setlength{\tabcolsep}{2pt}
\resizebox{0.8\linewidth}{!}{
\begin{tabular}{lccc}
\toprule[1pt]
     Method   & PSNR $\uparrow$ & SSIM $\uparrow$ & LPIPS $\downarrow$ \\ \midrule[1pt]
DBARF (Chen, 2023)    & 22.97  &   0.73  &   0.30                     \\
CoPoNeRF (Hong, 2024)  & 21.60  & 0.67  & 0.27                  \\
Ours    & \textbf{28.73}  &   \textbf{0.82}  &  0.29              \\ \bottomrule[1pt]
\end{tabular}
}
\vspace{-8pt}
\caption{\textbf{Comparison against generalizable NeRF methods on novel-view synthesis across all 8 scenes on the  Tanks and Temple dataset.}}
\label{table:nvc-compare}
\vspace{-15pt}
\end{table}

\begin{figure*}[h]
  \centering
\includegraphics[width=1.0\textwidth]{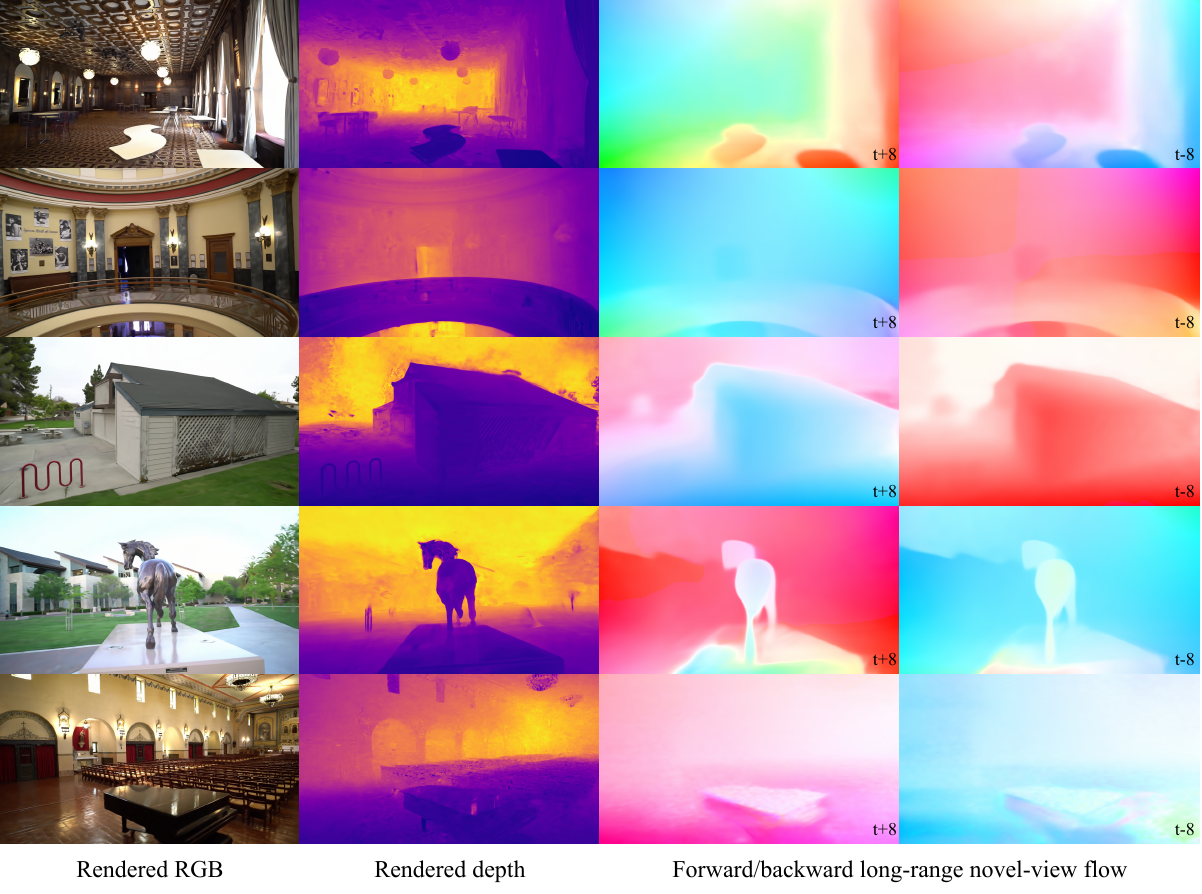}
  \vspace{-20pt}
  \caption{\textbf{Additional visualization results on long-range frame flow estimation on the Tanks and Temples dataset}. T+8 and t-8
denote forward and backward flow with a frame interval of 8, respectively.}
  \label{fig:tanks-flow}
\end{figure*}

\begin{table}[h]
\setlength{\tabcolsep}{2pt}
\resizebox{\linewidth}{!}{
\begin{tabular}{lccccccc}
\toprule[1pt]
     0079\_00   & Abs Rel $\downarrow$ & Sq Rel $\downarrow$ & RMSE $\downarrow$ & RMSE log $\downarrow$ & $\delta_1 \uparrow$ & $\delta_2 \uparrow$ & $\delta_3 \uparrow$ \\ \midrule[1pt]
BARF~\cite{lin2021barf}    & 0.208  &   0.165  &   0.588  &   0.263  &   0.639  &   0.896  &   0.983                  \\
NeRFmm~\cite{wang2021nerf}  &0.494 &  1.049 &  1.419 &  0.534 &  0.378 &  0.567 &  0.765                      \\
SC-NeRF~\cite{song2023sc} & 0.360  &   0.450  &   0.902  &   0.396  &   0.407  &   0.730  &   0.908 \\
Nope-NeRF~\cite{bian2023nope} & 0.099 & 0.047 & 0.335 & 0.128 & 0.904 & 0.995 & 1.000                       \\
Ours    & \textbf{0.040}  &   \textbf{0.006}  &   \textbf{0.106}  &   \textbf{0.057}  &   \textbf{0.993}  &   \textbf{1.000}  &   \textbf{1.000}                      \\ \bottomrule[1pt]
\end{tabular}
}
\vspace{-10pt}
\caption{\textbf{Depth map evaluation on ScanNet 0079\_00}.}
\vspace{-20pt}
\label{table:depth1}
\end{table}

\begin{table}[h]
\setlength{\tabcolsep}{2pt}
\resizebox{\linewidth}{!}{
\begin{tabular}{lccccccc}
\toprule[1pt]
     0418\_00   & Abs Rel $\downarrow$ & Sq Rel $\downarrow$ & RMSE $\downarrow$ & RMSE log $\downarrow$ & $\delta_1 \uparrow$ & $\delta_2 \uparrow$ & $\delta_3 \uparrow$ \\ \midrule[1pt]

BARF~\cite{lin2021barf}    & 0.718  &   1.715  &   1.563  &   0.630  &   0.205  &   0.569  &   0.769                    \\
NeRFmm~\cite{wang2021nerf}  &0.907  &   3.650  &   2.176  &   0.769  &   0.240  &   0.456  &   0.621                       \\
SC-NeRF~\cite{song2023sc} & 0.319  &   0.441  &   0.898  &   0.377  &   0.456  &   0.792  &   0.930 \\ 
Nope-NeRF~\cite{bian2023nope} & 0.152    & 0.137   & 0.645   & 0.185   & 0.738   &  0.998   & 0.997                       \\
Ours    & \textbf{0.034}  &   \textbf{0.010}  &   \textbf{0.118}  &   \textbf{0.070}  &   \textbf{0.984}  &   \textbf{0.995}  &   \textbf{0.998}                     \\ \bottomrule[1pt]
\end{tabular}
}
\vspace{-8pt}
\caption{\textbf{Depth map evaluation on ScanNet 0418\_00}.}
\label{table:depth2}
\vspace{-15pt}
\end{table}

\begin{table*}[h]
\resizebox{\textwidth}{!}{
\begin{tabular}{ccccclccclccclccclccc}
\toprule[1pt]
\multirow{2}{*}{scenes} && \multicolumn{3}{c}{Ours} &  & \multicolumn{3}{c}{Nope-NeRF~\cite{bian2023nope}} &  & \multicolumn{3}{c}{BARF~\cite{lin2021barf}} &  & \multicolumn{3}{c}{NeRFmm~\cite{wang2021nerf}} &  & \multicolumn{3}{c}{SC-NeRF~\cite{song2023sc}}               \\ \cline{3-5} \cline{7-9} \cline{11-13} \cline{15-17} \cline{19-21} 
                        & & $\text{RPE}_t\downarrow$    & $\text{RPE}_r\downarrow$    & ATE$\downarrow$    &  & $\text{RPE}_t$      & $\text{RPE}_r$      & ATE     &  & $\text{RPE}_t$    & $\text{RPE}_r$    & ATE    &  & $\text{RPE}_t$     & $\text{RPE}_r$     & ATE    &  & $\text{RPE}_t$                       & $\text{RPE}_r$   & ATE   \\ \midrule[1pt]
\multicolumn{1}{c}{\multirow{9}{*}{\rotatebox[origin=c]{90}{Tanks and Temple}}} &
Church                  & 0.035  & 0.093  & 0.008  &  & 0.034    & 0.008    & 0.008   &  & 0.114  & 0.038  & 0.052  &  & 0.626   & 0.127   & 0.065  &  & 0.836                     & 0.187 & 0.108 \\
\multicolumn{1}{c}{} & Barn                    & 0.076  & 0.162  & 0.008  &  & 0.046    & 0.032    & 0.004   &  & 0.314  & 0.265  & 0.050  &  & 1.629   & 0.494   & 0.159  &  & 1.317                     & 0.429 & 0.157 \\
\multicolumn{1}{c}{} & Museum                  & 0.125  & 0.187  & 0.007  &  & 0.207    & 0.202    & 0.020   &  & 3.442  & 1.128  & 0.263  &  & 4.134   & 1.051   & 0.346  &  & 8.339                     & 1.491 & 0.316 \\
\multicolumn{1}{c}{} & Family                  & 0.173  & 0.068  & 0.009  &  & 0.047    & 0.015    & 0.001   &  & 1.371  & 0.591  & 0.115  &  & 2.743   & 0.537   & 0.120  &  & 1.171                     & 0.499 & 0.142 \\
\multicolumn{1}{c}{} & Horse                   & 0.181  & 0.069  & 0.009  &  & 0.179    & 0.017    & 0.003   &  & 1.333  & 0.394  & 0.014  &  & 1.349   & 0.434   & 0.018  &  & 1.366                     & 0.438 & 0.019 \\
\multicolumn{1}{c}{} & Ballroom                & 0.115  & 0.101  & 0.008  &  & 0.041    & 0.018    & 0.002   &  & 0.531  & 0.228  & 0.018  &  & 0.449   & 0.177   & 0.031  &  & 0.328                     & 0.146 & 0.012 \\
\multicolumn{1}{c}{} & Francis                 & 0.177  & 0.296  & 0.023  &  & 0.057    & 0.009    & 0.005   &  & 1.321  & 0.558  & 0.082  &  & 1.647   & 0.618   & 0.207  &  & 1.233                     & 0.483 & 0.192 \\
\multicolumn{1}{c}{} & Ignatius                & 0.081  & 0.049  & 0.006  &  & 0.026    & 0.005    & 0.002   &  & 0.736  & 0.324  & 0.029  &  & 1.302   & 0.379   & 0.041  &  & 0.533                     & 0.240 & 0.085 \\ 
\multicolumn{1}{c}{} & mean                    & 0.120  & 0.128  & 0.010  &  & 0.080    & 0.038    & 0.006   &  & 1.046  & 0.441  & 0.078  &  & 1.735   & 0.477   & 0.123  &  & \multicolumn{1}{l}{1.890} & 0.489 & 0.129 \\ \bottomrule[1pt]
\end{tabular}
}
\caption{\textbf{Camera pose comparison on the Tanks and Temples dataset}.}
\label{table:5}
\vspace{-10pt}
\end{table*}

\begin{figure*}[h]
  \centering
\includegraphics[width=1.0\textwidth]{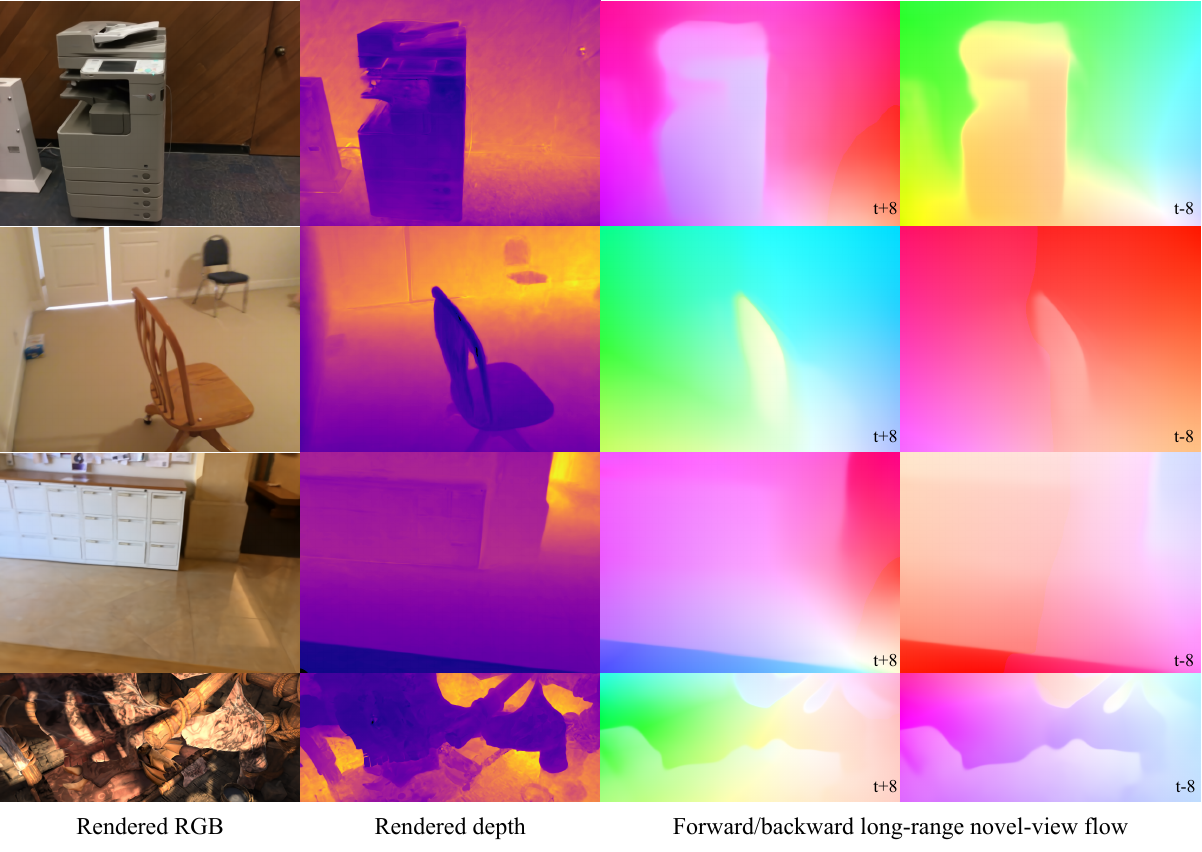}
  \vspace{-20pt}
  \caption{\textbf{Visualization on long-range frame flow estimation on the ScanNet and Sintel dataset}. T+8 and t-8 denote forward and backward flow with a frame interval of 8, respectively.}
  \label{fig:scannet-flow}
  \vspace{-15pt}
\end{figure*}

\section{Comparison with other SOTA methods} 
We compare the novel-view synthesis and depth estimation across all 4 scenes on ScanNet, as shown in Tab.~\ref{table:depth_compare}. Our method is clearly better than LocalRF~\cite{meuleman2023progressively} and CF-3DGS~\cite{fu2023colmap}. Since both LocalRF and CF-3DGS employ an incremental manner, they lack global bundle adjustment to correct the depth scale discrepancy among each sub-model/local Gaussian, leading to inconsistent geometry. 

We also compare the novel-view synthesis against generalizable NeRF methods DBARF~\cite{chen2023dbarf} and CoPoNeRF~\cite{hong2023unifying} across all 8 scenes on the Tanks and Temple dataset, as shown in Tab.~\ref{table:nvc-compare}. For CoPoNeRF designed for two-view geometry, we select the $i-1$ and $i+1$ frame as the context frames, and query the test frame $i$ in the middle. We largely outperform generalizable NeRF although they have been pretrained on large-scale datasets.

\section{Drastic camera motion scenes and pose visualization} 
\begin{figure}[!t]
  \centering
  \includegraphics[width=1.0\linewidth]{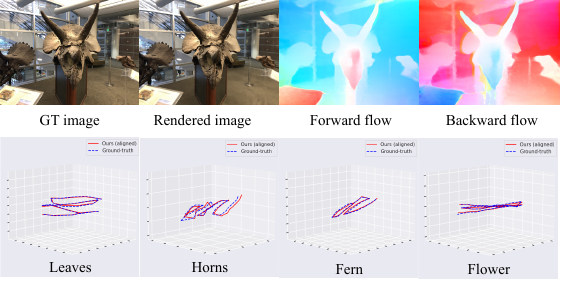}
  \vspace{-20pt}
   \caption{\textbf{Visualization results on the LLFF dataset.} Our method can handle large and fast camera motion, and render out plausible flow.}
   \label{fig:LLFF}
   \vspace{-20pt}
\end{figure}

\begin{figure}[!t]
  \centering
  \includegraphics[width=1.0\linewidth]{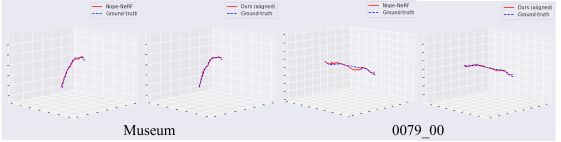}
  \vspace{-5pt}
   \caption{\textbf{Visualization of camera pose comparson against Nope-NeRF on the Museum scene and 0079$\_$00 scene.} Our method can estimate more accurate camera poses when large camera rotation exists. Better viewed when zoom in.}
   \label{fig:pose-compare}
   \vspace{-10pt}
\end{figure}

We test our method on several scenes of the LLFF dataset which contains irregular and fast camera motion. The visualization of the rendered flow and camera pose can be found in Fig.~\ref{fig:LLFF}. The visualization results show that our method can render plausible flow and estimate camera poses even under drastic camera motion. We also provide the visualization comparison (see Fig.~\ref{fig:pose-compare}) on 2 challenging scenes, from which the Museum scene has the maximum rotation of 76.2, and the 0079\_00 has the maximum rotation of 54.4. Our method performs better than Nope-NeRF when the camera rotation is large. 

\section{Additional results}
\begin{figure*}[h]
  \centering
\includegraphics[width=1.0\textwidth]{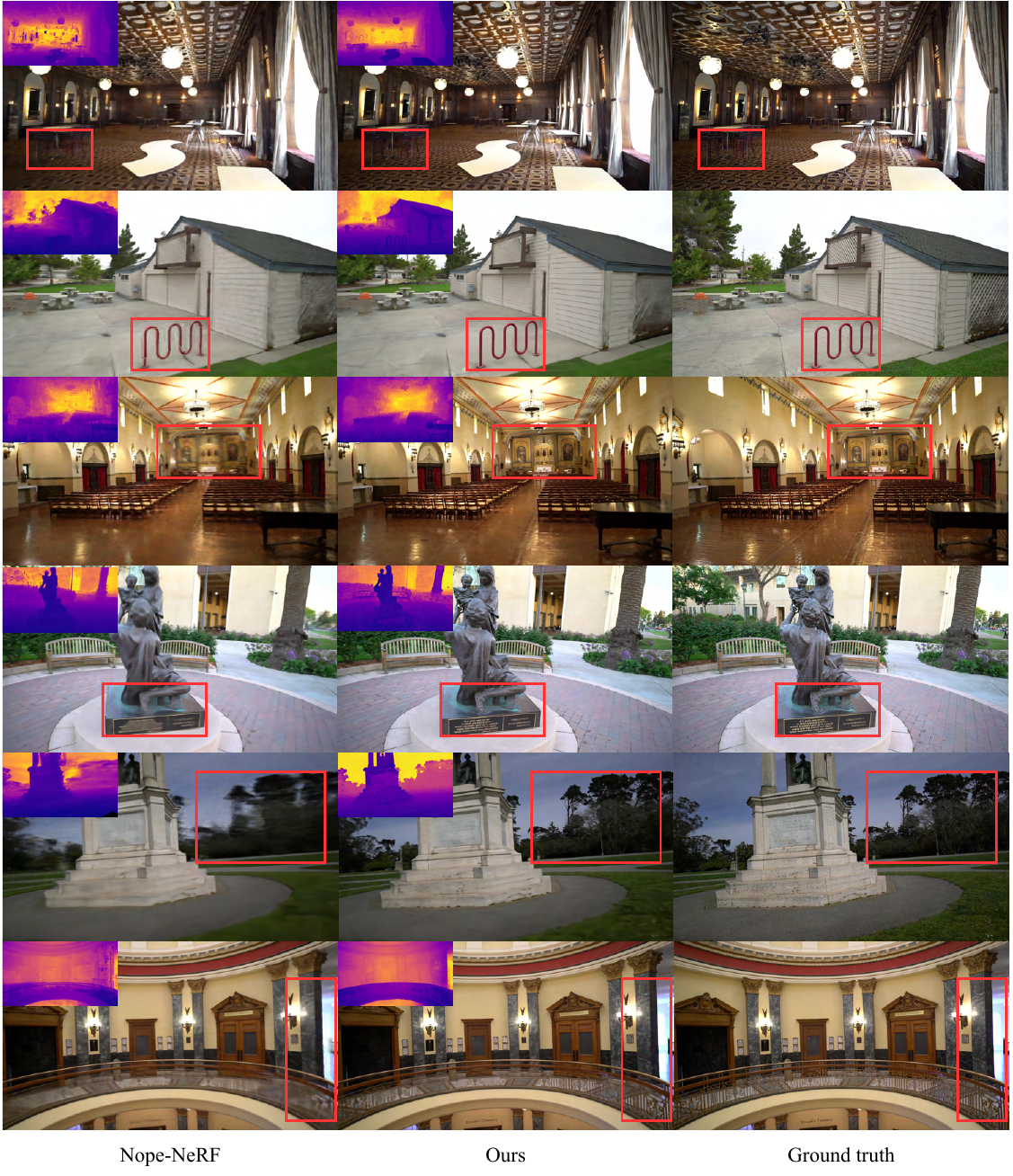}
  \vspace{-20pt}
  \caption{\textbf{Qualitative novel-view synthesis and depth estimation results on Tanks and Temples}. Compared with Nope-NeRF~\cite{bian2023nope}, our method can produce more photo-realistic novel-view synthesis results, and yield smoother depth maps while preserving more structure details. Better viewed when zoomed in.}
  \label{fig:tanks-nvs}
\end{figure*}

We provide several additional visualization results for novel-view synthesis, novel-view depth, and long-range novel-view flow predictions across several scenes, as shown in Fig.~\ref{fig:tanks-flow} and Fig.~\ref{fig:scannet-flow}. The qualitative results indicate consistent predictions among the novel-view images, novel-view depths, and novel-view flows, demonstrating that all our optimization objectives are indeed coupled. We also present a further visual comparison of the novel-view images and depth predictions on the Tanks and Temples dataset, as shown in Fig.~\ref{fig:tanks-nvs}. Compared to the state-of-art method Nope-NeRF~\cite{bian2023nope}, our method produces more photo-realistic novel-view images and significantly smoother depth maps with fewer artifacts, clearly validating the effectiveness of the proposed flow-enhanced novel-view synthesis and flow-enhanced geometry. 

\begin{figure*}[h]
  \centering
\includegraphics[width=0.97\textwidth]{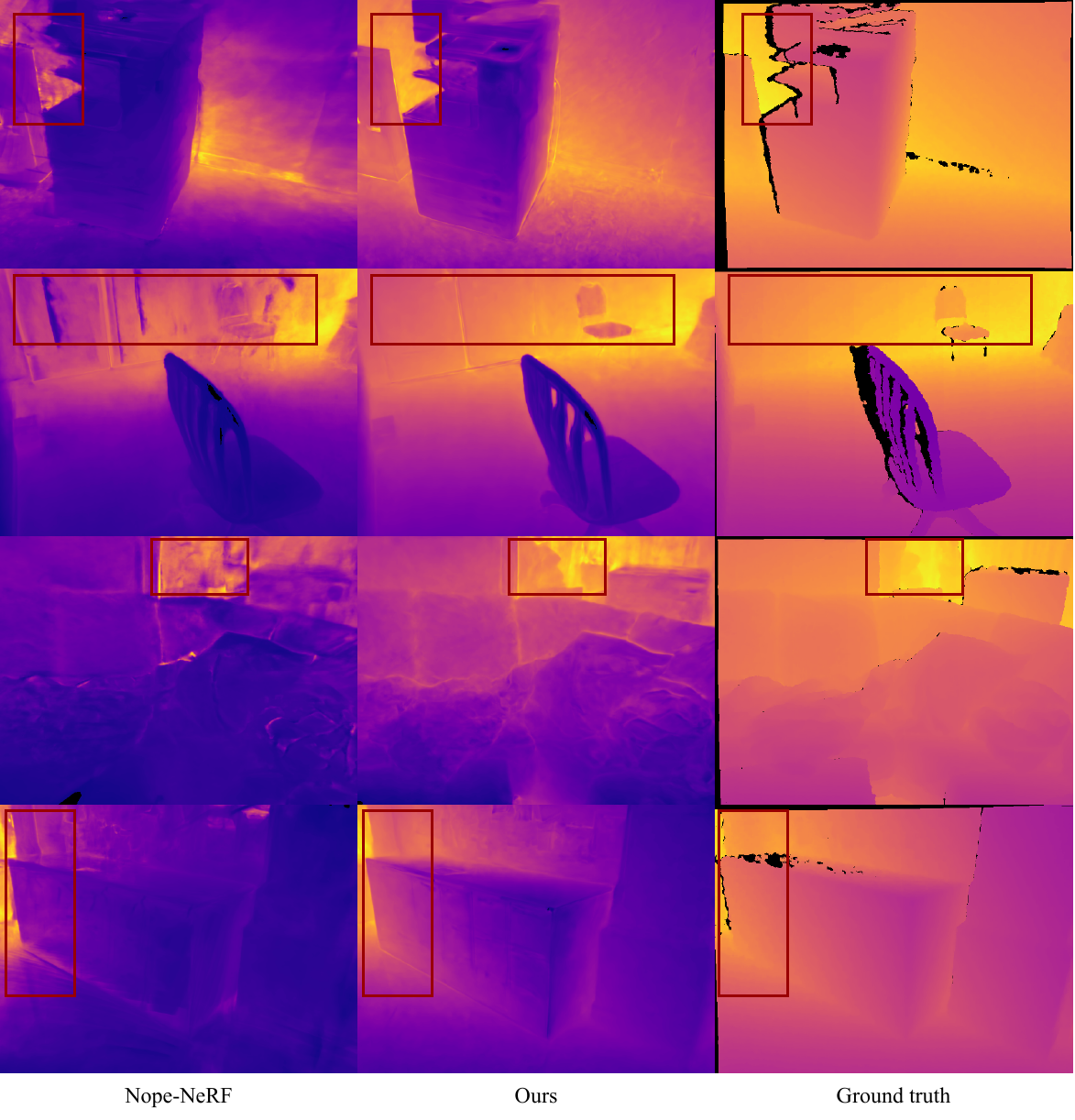}
  \vspace{-10pt}
  \caption{\textbf{Qualitative depth prediction comparison on ScanNet}. Compared with Nope-NeRF~\cite{bian2023nope}, our method predicts smoother novel-view depth with much fewer artifacts and preserves more structure details.}
  \label{fig:scannet-depth}
  \vspace{-10pt}
\end{figure*}

Besides, we provide the per-scene depth prediction results on the ScanNet dataset. The qualitative results are displayed in Fig.~\ref{fig:scannet-depth}, while the quantitative results are shown in Tab.~\ref{table:depth1},~\ref{table:depth2}, ~\ref{table:depth3} and~\ref{table:depth4}. Both the qualitative and quantitative results demonstrate that our method predicts depth maps significantly better than all other methods.

\begin{figure*}[h]
  \centering
\includegraphics[width=1.0\textwidth]{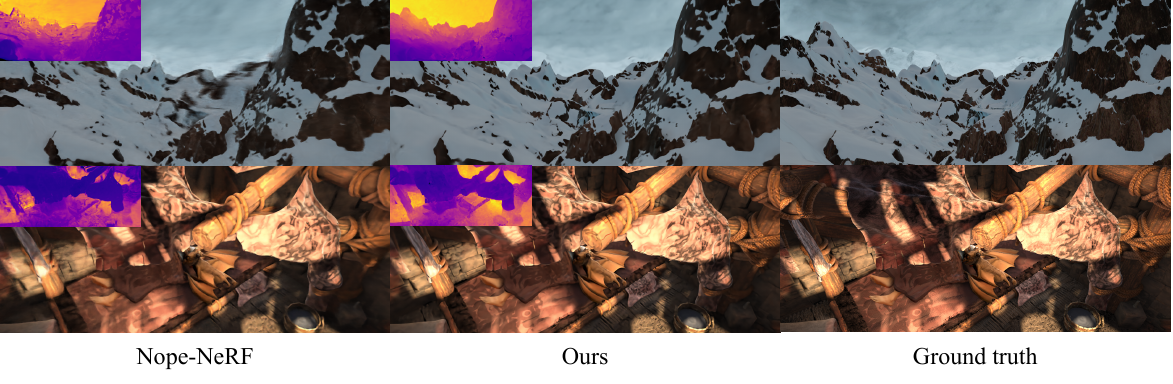}
  \vspace{-5pt}
  \caption{\textbf{Novel view synthesis and depth prediction comparison on Sintel}. Compared with Nope-NeRF~\cite{bian2023nope}, our method produces more photo-realistic novel-view images and depth maps with sharper details.}
  \label{fig:Sintel}
\end{figure*}

We also present additional novel-view synthesis and pose estimation results on the Sintel dataset (see Tab.~\ref{table:6} and Tab.~\ref{table:7}). Note that the Sintel dataset contains high-resolution images and large camera motion, and our method significantly outperforms Nope-NeRF in both tasks. We also provide the visualization of the depth prediction and novel view synthesis on Sintel (see Fig.~\ref{fig:Sintel}). Our method can generate depth maps with sharper details and produce more photo-realistic novel-view images.

\begin{table}[h]
\setlength{\tabcolsep}{2pt}
\resizebox{\linewidth}{!}{
\begin{tabular}{lccccccc}
\toprule[1pt]
     0301\_00   & Abs Rel $\downarrow$ & Sq Rel $\downarrow$ & RMSE $\downarrow$ & RMSE log $\downarrow$ & $\delta_1 \uparrow$ & $\delta_2 \uparrow$ & $\delta_3 \uparrow$ \\ \midrule[1pt]
BARF~\cite{lin2021barf}    & 0.179  &   0.146  &   0.502  &   0.268  &   0.736  &   0.883  &   0.938                   \\
NeRFmm~\cite{wang2021nerf}  &0.444  &   0.830  &   1.239  &   0.481  &   0.397  &   0.680  &   0.845                     \\
SC-NeRF~\cite{song2023sc} & 0.383  &   0.378  &   0.810  &   0.452  &   0.360  &   0.663  &   0.846 \\
Nope-NeRF~\cite{bian2023nope} & 0.185    & 0.252   & 0.711   & 0.233   & 0.792   &  0.918   & 0.958                       \\
Ours    & \textbf{0.036}  &   \textbf{0.006}  &   \textbf{0.127}  &   \textbf{0.053}  &   \textbf{0.991}  &   \textbf{1.000}  &   \textbf{1.000}                     \\ \bottomrule[1pt]
\end{tabular}
}
\vspace{-8pt}
\caption{\textbf{Depth map evaluation on ScanNet 0301\_00}.}
\label{table:depth3}
\vspace{-8pt}
\end{table}

\begin{table}[h]
\setlength{\tabcolsep}{2pt}
\resizebox{\linewidth}{!}{
\begin{tabular}{lccccccc}
\toprule[1pt]
     0431\_00   & Abs Rel $\downarrow$ & Sq Rel $\downarrow$ & RMSE $\downarrow$ & RMSE log $\downarrow$ & $\delta_1 \uparrow$ & $\delta_2 \uparrow$ & $\delta_3 \uparrow$ \\ \midrule[1pt]
BARF~\cite{lin2021barf}    & 0.398  &   0.710  &   1.307  &   0.444  &   0.381  &   0.655  &   0.847                  \\
NeRFmm~\cite{wang2021nerf}  &0.514  &   1.354  &   1.855  &   0.562  &   0.250  &   0.539  &   0.742                       \\
SC-NeRF~\cite{song2023sc} & 0.608  &   1.300  &   1.706  &   0.677  &   0.225  &   0.446  &   0.645  \\
Nope-NeRF~\cite{bian2023nope} & 0.127    & 0.111   & 0.579   & 0.160   & 0.877   &  0.978   & 0.994                       \\
Ours    & \textbf{0.078}  &   \textbf{0.028}  &   \textbf{0.251}  &   \textbf{0.107}  &   \textbf{0.960}  &   \textbf{0.978}  &   \textbf{0.999}                     \\ \bottomrule[1pt]
\end{tabular}
}
\vspace{-8pt}
\caption{\textbf{Depth map evaluation on ScanNet 0431\_00}.}
\label{table:depth4}
\vspace{-8pt}
\end{table}

\begin{table}[h]
\setlength{\tabcolsep}{2pt}
\resizebox{\linewidth}{!}{
\begin{tabular}{cccclccc}
\toprule[1pt]
\multirow{2}{*}{Method} & \multicolumn{3}{c}{mountain1}             &  & \multicolumn{3}{c}{sleeping2}                        \\ \cline{2-4} \cline{6-8} 
                        & PSNR$\uparrow$           & SSIM$\uparrow$          & LPIPS$\downarrow$         &  & PSNR           & SSIM          & LPIPS              \\ \midrule[1pt]
Nope-NeRF~\cite{bian2023nope}               & 27.24 & 0.86          & 0.33          &  & 27.93          & 0.79 & 0.40      \\
Ours                    & \textbf{31.24}          & \textbf{0.93} & \textbf{0.24} &  & \textbf{31.19} & \textbf{0.88} & \textbf{0.32} \\ \bottomrule[1pt]
\end{tabular}
}
\caption{\textbf{Novel view synthesis comparison on Sintel}.}
\label{table:6}
\vspace{-15pt}
\end{table}

\begin{table}[h]
\setlength{\tabcolsep}{2pt}
\resizebox{\linewidth}{!}{
\begin{tabular}{cccclccc}
\toprule[1pt]
\multirow{2}{*}{Method} & \multicolumn{3}{c}{mountain1}             &  & \multicolumn{3}{c}{sleeping2}                        \\ \cline{2-4} \cline{6-8} 
                        & $\text{RPE}_t\downarrow$           & $\text{RPE}_r\downarrow$          & ATE$\downarrow$         &  & $\text{RPE}_t$           & $\text{RPE}_r$          & ATE              \\ \midrule[1pt]
Nope-NeRF~\cite{bian2023nope}               & 1895 & 0.237          & 21.57          &  & 3.616          & 0.574 & 0.005      \\
Ours                    & \textbf{1332}          & 0.248 & \textbf{3.501} &  & \textbf{3.566} & \textbf{0.537} & \textbf{0.002} \\ \bottomrule[1pt]
\end{tabular}
}
\caption{\textbf{Camera pose comparison on Sintel}.}
\label{table:7}
\end{table}

\section{Limitations and future work}
Given accurate pixel-wise and frame-wise correspondence prediction, one can easily compute the relative poses for frame pairs using either an analytical pose-solver or performing a motion-only bundle adjustment. We have not yet explored the potential of the predicted novel-view flow in this promising direction.

\clearpage


\end{document}